\documentclass[lettersize, journal, twoside]{IEEEtran}

\usepackage{amsmath}
\usepackage{amsfonts}
\usepackage{amssymb}

\usepackage{amsthm}
\usepackage{mathtools}
\usepackage{tabularx,booktabs}
\usepackage{graphicx}
\usepackage{subfigure}
\usepackage{enumerate}
\usepackage{float}
\usepackage{url}
\usepackage{verbatim}
\usepackage{cite}
\usepackage{diagbox}
\usepackage{multirow}
\usepackage{makecell}
\usepackage{algorithm}  
\usepackage{algorithmicx}  
\usepackage{algpseudocode} 
\usepackage{bbding}
\usepackage{bm}
\usepackage{cases}
\usepackage[linkcolor=black,citecolor=black,urlcolor=black,colorlinks=true]{hyperref}
\usepackage{tikz}
\usepackage{etoolbox}
\usepackage{graphicx}
\usepackage{pifont}
\usepackage{threeparttable}
\usepackage{graphicx}
\usepackage{subcaption}

\hypersetup{colorlinks=true, urlcolor=cyan}
\usepackage{xcolor}
\DeclareRobustCommand{\add}[1]{%
  {\textcolor{black}{#1}}%
}

\DeclareRobustCommand{\addagain}[1]{%
  {\textcolor{black}{#1}}%
}
\graphicspath{{./figures/}}
\DeclareGraphicsExtensions{.png,.jpg,.eps,.pdf}

\begin{document}

\title{\LARGE \bf
A Hierarchical Onboard Vision-Language-Navigation Framework \\ for Aerial Robots in Complex Environments
}

\author{Yuze Wu\textsuperscript{1,2,$\ast$}, Mo Zhu\textsuperscript{1,2,$\ast$}, Xingxing Li\textsuperscript{2,$\ast$}, Yuheng Du\textsuperscript{2}, Zhaoqi Wang\textsuperscript{2}, Yuxin Fan\textsuperscript{1,2}, \\ Wenjun Li\textsuperscript{1,2}, Kenghou Hoi\textsuperscript{1,2}, Zhichao Han\textsuperscript{1,2}, Xin Zhou\textsuperscript{1,2,$\dagger$}, and Fei Gao\textsuperscript{1,2, $\dagger$}
	\thanks{This work was supported in part by the National Key R\&D Program of China under grant no. 2023YFB4706600, in part by the Fundamental Research Funds for the Central Universities under Project No. 226-2025-00114, and in part by the National Natural Science Foundation of China under Grant 62322314. } 
	\thanks{$^\ast$These authors contributed equally to this work.}
    \thanks{$^\dagger$Corresponding author. E-mail: iszhouxin@zju.edu.cn (Xin Zhou), fgaoaa@zju.edu.cn (Fei Gao).}
	\thanks{\textsuperscript{1}State Key Laboratory of Industrial Control Technology, Zhejiang University, Hangzhou 310027, China.}
	\thanks{\textsuperscript{2}Differential Robotics, Huzhou 310027, China.}	
    \thanks{Manuscript received May 21, 2026; accepted November 9, 2026.}
}

\markboth{IEEE Transactions on Robotics. Preprint Version. Accepted November 9, 2026}{Wu \MakeLowercase{\textit{et al.}}: A Hierarchical Onboard Vision-Language-Navigation Framework for Aerial Robots in Complex Environments}



\setcounter{figure}{-2}
	\makeatletter
	\let\@oldmaketitle\@maketitle
	\renewcommand{\@maketitle}{\@oldmaketitle
		\begin{center}
			\includegraphics[width=0.88\linewidth]{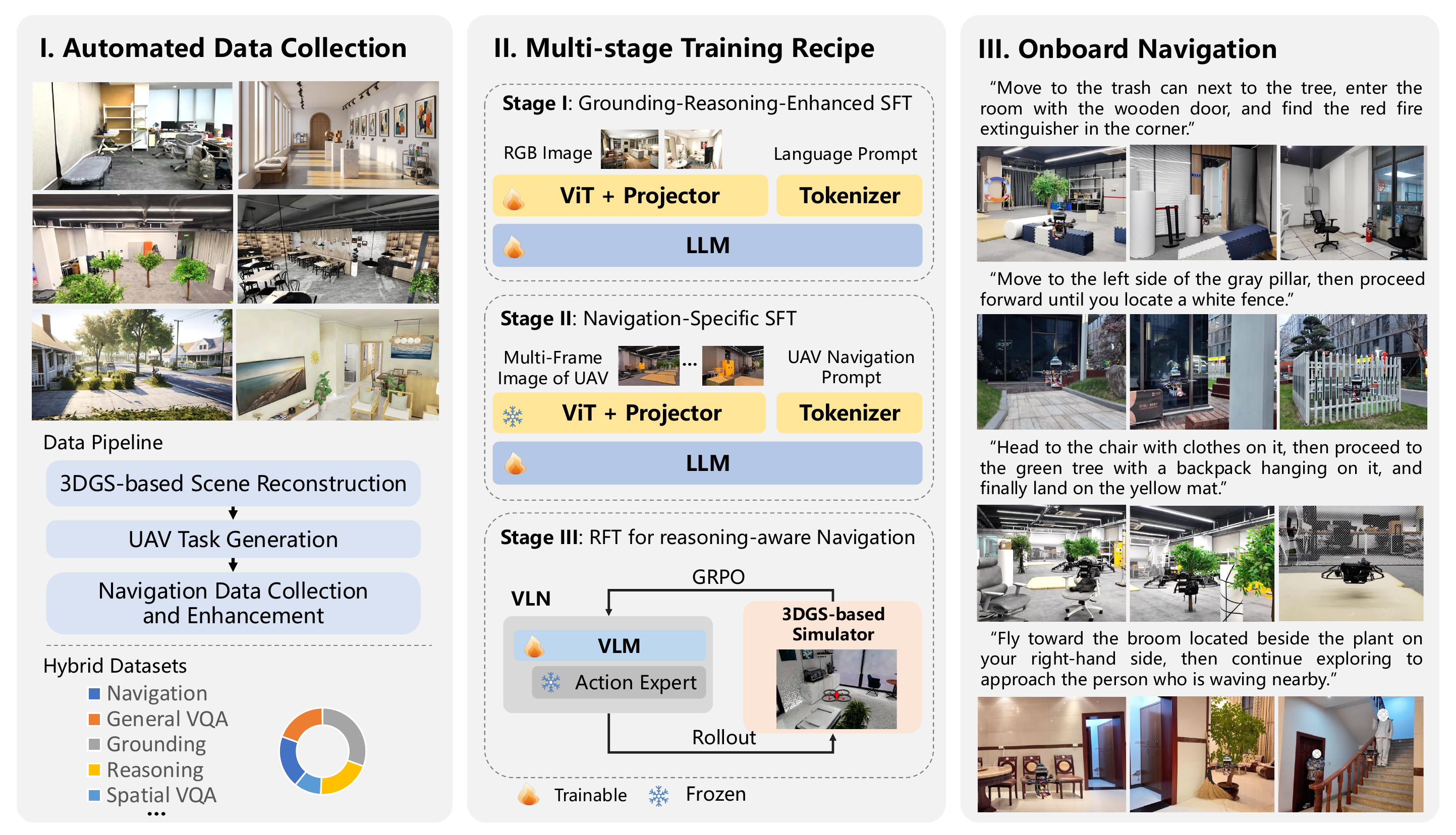}
            \captionof{figure}{
            \label{fig:overview}
            Overview of the dataset, training, and deployment of the proposed VLN-AR. (I) We design a fully automated aerial navigation data collection pipeline and construct a large-scale 3DGS-based dataset. (II) We introduce a progressive three-stage training recipe to sequentially strengthen spatial grounding, scene reasoning, and long-horizon navigation. (III) VLN-AR generates efficient, fast, and collision-free navigation commands onboard under language instructions and visual observations.
            \vspace{-0.7cm}
            }
		\end{center}
	}
	\makeatother
	\maketitle
	\setcounter{figure}{1}

\begin{abstract}
    Unmanned Aerial Vehicles (UAVs) are expected to achieve semantic reasoning, intelligent decision-making, and long-horizon navigation under strict onboard constraints. While recent foundation-model-driven approaches have shown promising progress toward this goal, existing methods remain limited by challenges in 3D data availability, trajectory feasibility, long-horizon reasoning, and computational efficiency in complex environments.
    To bridge these gaps, we propose VLN-AR, a hierarchical vision-language-navigation (VLN) framework for aerial robots. VLN-AR decouples multimodal reasoning and low-level control, enabling structured prediction of 3D pose commands and navigation intentions that are translated into dynamically feasible actions. To mitigate data scarcity and domain gaps, we introduce a fully automated 3D Gaussian Splatting (3DGS)-based pipeline to generate scalable photorealistic UAV navigation data. A progressive three-stage training strategy further enhances grounding, scene reasoning, and complex navigation. In addition, a lightweight cognitive agent improves long-horizon reasoning via memory-aware task decomposition. Finally, we design onboard optimization techniques, including token sparsification and model lightweighting, achieving 7.1×–9.4× speedup while maintaining high navigation performance on resource-limited UAV platforms. Experiments show great improvements over existing methods in both simulation and real-world UAV navigation. Project website: \url{https://wyz000.github.io/VLN-AR/}.
\end{abstract}


\section{Introduction}
\IEEEPARstart{A}{utonomous} unmanned aerial vehicles (UAVs) have been widely deployed in diverse applications including infrastructure inspection, search and rescue, environmental monitoring, and logistics delivery. In such scenarios, UAVs are often required to operate in unstructured and partially observable environments. Beyond reliable low-level flight control, these settings demand high-level intelligence, including the ability to comprehend natural-language instructions, reason about spatial semantics, and generate safe and dynamically feasible actions under strict onboard computational constraints. Enabling such capabilities is critical to advancing UAV systems from task-specific navigation toward general embodied autonomy. While Vision-Language-Navigation (VLN) \cite{zhang2025embodiednavigationfoundationmodel} and Vision-Language-Action (VLA) \cite{kim2024openvlaopensourcevisionlanguageactionmodel, black2024pi0visionlanguageactionflowmodel} paradigms have shown considerable potential for autonomous robots, such methods remain underexplored for agile, resource-constrained UAVs.
    
    Most existing UAV navigation methods follow modular perception–planning–control pipelines \cite{9811688,9981489,10044964}, relying on predefined algorithms, hand-crafted policies, or explicit mapping strategies. While effective in well-defined tasks, they lack general task understanding and reasoning capabilities, particularly the ability to interpret open-vocabulary language instructions. 
    Recent advances in multimodal large language models (MLLMs) \cite{liu2023visualinstructiontuning,openai2023gpt4,team2023gemini,bai2023qwenvl} have enabled VLN and VLA systems for ground robots, allowing natural language to be grounded in visual observations and translated into executable commands within a unified framework. However, adapting these methods to agile aerial robots is non-trivial due to fundamental differences in system dynamics, action spaces, data distributions, safety constraints, and deployment requirements. In particular, several critical challenges are as follows:

    \begin{itemize}
        \item \textbf{Scarcity of 3D navigation data and domain shift.} Most navigation foundation models are pretrained on large-scale ground-view 2D navigation datasets, which differ drastically from UAV-centric views that exhibit large variations in pose and scale. Existing UAV navigation datasets are often limited to outdoor environments with coarse visual features and insufficient fine-grained structural details, resulting in sim-to-real domain gaps.
        
        \item \textbf{Mismatch between 3D action representation and safety-aware control.} UAV navigation requires safe and reliable control. End-to-end continuous commands derived from foundation models often fail to guarantee safety or satisfy dynamical constraints, while semantically discrete action representations offer limited maneuverability. Balancing 3D maneuverability, control stability, and collision avoidance remains a core challenge.
        
        \item \textbf{Insufficient reasoning for long-horizon embodied navigation.} Existing VLN/VLA methods typically rely on single-frame or short-term reasoning, making it difficult to integrate historical observations in complex indoor environments. Long-horizon aerial navigation requires spatial grounding, conditional inference, and sequential task decomposition, which cannot be sufficiently supported by single-step prediction schemes.
        
        \item \textbf{Prohibitive onboard computational constraints.} State-of-the-art foundation models are computationally intensive, making real-time deployment on small UAVs with strict size, weight, and power (SWaP) constraints challenging \cite{ravichandran2025distillingondevicelanguagemodels}. Practical onboard deployment therefore requires preserving strong inference capability while significantly reducing computational and memory overhead.
    \end{itemize}

    To address these challenges, we propose \textbf{VLN-AR}, a hierarchical \textbf{V}ision \textbf{L}anguage \textbf{N}avigation framework for \textbf{A}erial \textbf{R}obots in complex environments. The proposed system integrates data generation, multimodal learning, and onboard deployment into a unified closed-loop framework (Fig.~\ref{fig:overview}). At its core, VLN-AR decomposes the navigation problem into two tightly coupled components: a navigation-oriented vision language model (VLM) that performs multimodal reasoning and predicts structured navigation intentions, and a lightweight action expert that translates these intentions into dynamically feasible control commands. Instead of directly generating low-level actions, the model generates diverse navigation representations tailored to different task requirements. Specifically, it outputs a structured representation comprising 3D pose commands, 2D bounding-box guidance, and chain-of-thought (CoT) reasoning, establishing an effective interface between high-level cognition and low-level control. The proposed model achieves a maximum success rate of 98.1\% on single navigation tasks and an overall aerial navigation success rate of 92.7\%, \add{while attaining success rates of 78.15\% and 76.88\% on the AerialVLN and OpenFly datasets, respectively}.

    To mitigate flight data scarcity and sim-to-real domain gaps in UAV navigation, we develop a high-fidelity data generation pipeline based on 3D Gaussian Splatting (3DGS) \cite{kerbl2023gaussiansplatting}, \add{enabling photorealistic and temporally consistent UAV-centric observations. Combined with an automated data collection and annotation pipeline, it produces} a large-scale 3D aerial navigation dataset with approximately 4 million samples, covering navigation data, general visual question answering (VQA), grounding data, reasoning data, spatial VQA, and more.

    Building upon this data foundation, we introduce a progressive three-stage training strategy that sequentially enhances multimodal grounding, navigation-specific motion prediction, and reasoning-aware navigation, as shown in Fig.~\ref{fig:overview}(II). The first two stages perform supervised fine-tuning (SFT) on large-scale mixed datasets to improve scene understanding and spatial reasoning for navigation, while the third stage  \add{applies reinforcement learning in closed-loop simulation to further improve reasoning-aware navigation} \cite{shao2024deepseekmathpushinglimitsmathematical}.
    
    To support long-horizon and ambiguity-aware navigation, we incorporate a lightweight cognitive-driven agent that aggregates historical observations, task states, and action states into a unified temporal memory, enabling structured reasoning, task decomposition, intelligent exploration, and failure recovery in partially observable environments.  \add{For efficient onboard deployment, we further combine instruction-guided visual token sparsification with operator-level acceleration and low-precision computation, achieving} a 7.1×--9.4× speedup on resource-limited UAV platforms.

    \add{This work substantially extends the previous work~\cite{wu2025vla}, which introduces preliminary components of the 3DGS-based data pipeline, hierarchical navigation architecture, training framework, and onboard deployment stack. The present study further introduces a cognitive navigation agent, instruction-guided token sparsification, and richer structured model outputs for planning and replanning. Additionally, the training and robustness evaluation have been significantly expanded, and a wider range of scenarios have been tested, thereby constructing a more complete onboard aerial VLN framework.} The main contributions are summarized as follows:
    \begin{itemize}
        \item [1)] We propose VLN-AR, a hierarchical onboard VLN framework that \add{integrates high-level task reasoning with diverse action representations to} enable \add{efficient,} closed-loop semantic aerial navigation in complex 3D environments.
        \item [2)] We present a cognitive navigation agent with dynamic memory fusion and heuristic task decomposition to enhance long-horizon reasoning, resolve ambiguous instructions, and improve resilience under partial observability in unknown environments.
        \item [3)] We develop systematic onboard optimization techniques, including instruction-guided visual token sparsification and model lightweighting, achieving a 7.1×–9.4× inference speedup while maintaining high navigation performance on resource-limited UAV platforms.
        \item [4)] We conduct comprehensive evaluations in both simulation and real-world environments, demonstrating that the proposed VLN-AR achieves state-of-the-art performance in terms of success rate, accuracy, robustness, and computational efficiency.
    \end{itemize}

\section{Related Works}
    \label{sec:Realated_Works}
    
    \subsection{Traditional Modular UAV Navigation}
    Early autonomous navigation systems for UAVs predominantly adopt modular, cascaded architectures following the classic perception--decision--planning--control pipeline \cite{9102390,10816079,wu2025shapeadaptiveplanningcontroldeformable,9691888}. State estimation and environmental perception are typically implemented via LiDAR-based or visual perception modules \cite{8421746, 9372856}, while semantic scene understanding is achieved through object detectors such as YOLO \cite{wang2024yolov10}, Faster R-CNN \cite{ren2015faster}, and 3D scene graphs \cite{chen2025irsinstancelevel3dscene}. High-level mission logic is subsequently resolved by rule-based or decision-tree modules, followed by motion planning algorithms such as A* \cite{hart1968formal}, RRT \cite{karaman2011sampling}, Fast-Planner \cite{8758904}, and Ego-Planner \cite{9309347} to generate feasible reference trajectories. 
    
    Such conventional approaches offer clear interpretability, robust stability, and low computational overhead. However, they are often customized for specific application scenarios, such as target tracking \cite{9561948}, gap traversal \cite{wu2026precise}, and autonomous landing \cite{10328688}, and lack the ability to understand open-vocabulary language instructions or exploit rich semantic information from visual observations. Furthermore, the sequential multi-module structure is prone to error propagation and accumulation, leading to significant performance degradation in cluttered, dynamic, and unstructured environments.  

    \vspace{-0.05cm}
    \subsection{Navigation Foundation Models for UAVs}
    \label{sec:Nav_model}
\begin{figure}[t]
      \centering
      \includegraphics[width=0.82\linewidth]{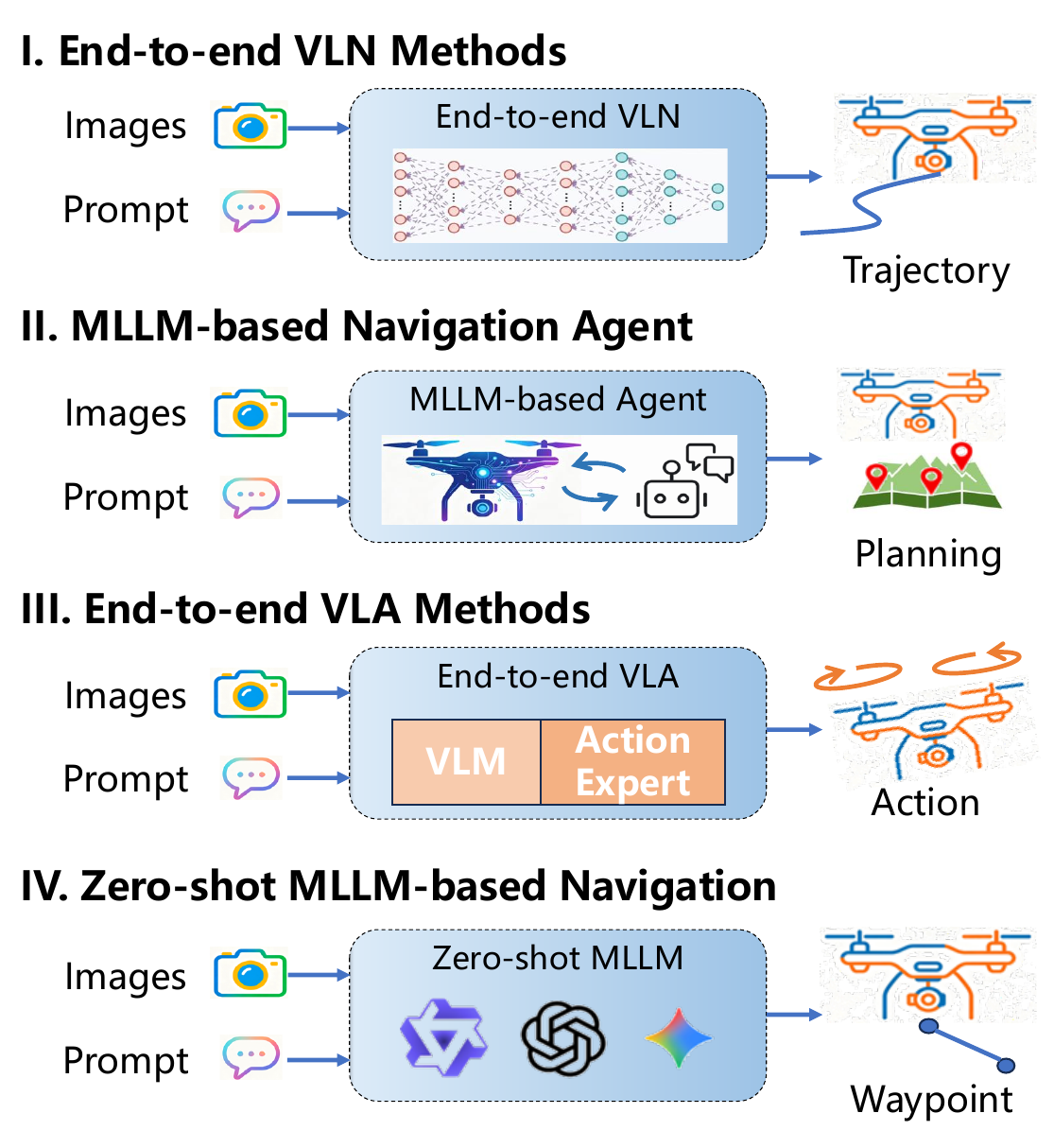}
      \caption{Different UAV navigation foundation models.
      }
      \label{fig:vla_method}
      \vspace{-0.3cm}
    \end{figure}

    We categorize existing UAV navigation foundation models into four representative groups based on their modeling paradigms (Fig.~\ref{fig:vla_method}). 
    (1) End-to-end VLN methods such as OpenFly \cite{gao2025openflycomprehensiveplatformaerial}, AerialVLN \cite{liu2023aerialvlnvisionandlanguagenavigationuavs}, OpenUAV \cite{wang2024realisticuavvisionlanguagenavigation}, and UAV-Flow \cite{wang2025uavflowcolosseorealworldbenchmark} learn direct mappings from visual inputs to flight trajectories; 
    (2) MLLM-based navigation agents, including CityNavAgent \cite{zhang2025citynavagent}\add{,} SPF \cite{hu2025seepointflylearningfree}\add{, SINGER \cite{adang2025singer}, and VLFly \cite{zhang2025grounded},} rely on pretrained \add{foundation} models for semantic grounding and reasoning, and convert detected targets into goal waypoints via rule-based modules. \add{However, these methods often rely on predefined goal representations or external planning modules, limiting their capability in complex open-world tasks.} 
    (3) End-to-end VLA methods such as OpenVLA \cite{kim2024openvlaopensourcevisionlanguageactionmodel}, $\pi_{0}$ \cite{black2024pi0visionlanguageactionflowmodel}, and GR00T N1.5 \cite{nvidia2025groot} are general embodied models that, although not designed for UAV navigation, exhibit strong transferability and are fine-tuned on navigation datasets \cite{wang2025uavflowcolosseorealworldbenchmark, chen2026towards, flymirage}. \add{Recent UAV-oriented VLA frameworks,such as GRaD-Nav++ \cite{chen2025grad} have utilized 3DGS-based data for training and onboard deployment. Simulators like GaussGym \cite{escontrela2025gaussgym} can provide more realistic visual data for such methods.}
    (4) Zero-shot MLLM-based navigation directly leverages pretrained multimodal foundation models to infer 3D waypoints and flight commands from language instructions, enabling strong generalization without task-specific training.
    
    Despite these advances, most existing methods still suffer from limitations in UAV-centric data coverage, temporal modeling, and real-time adaptability. In addition, the lack of explicit temporal reasoning makes them vulnerable to occlusions and long-horizon tasks, while reliance on prebuilt maps \cite{wang2024realisticuavvisionlanguagenavigation} further limits deployment in unknown environments and reduces robustness under dynamic scenes.

    \begin{figure*}[t]
      \centering
      \includegraphics[width=0.88\linewidth]{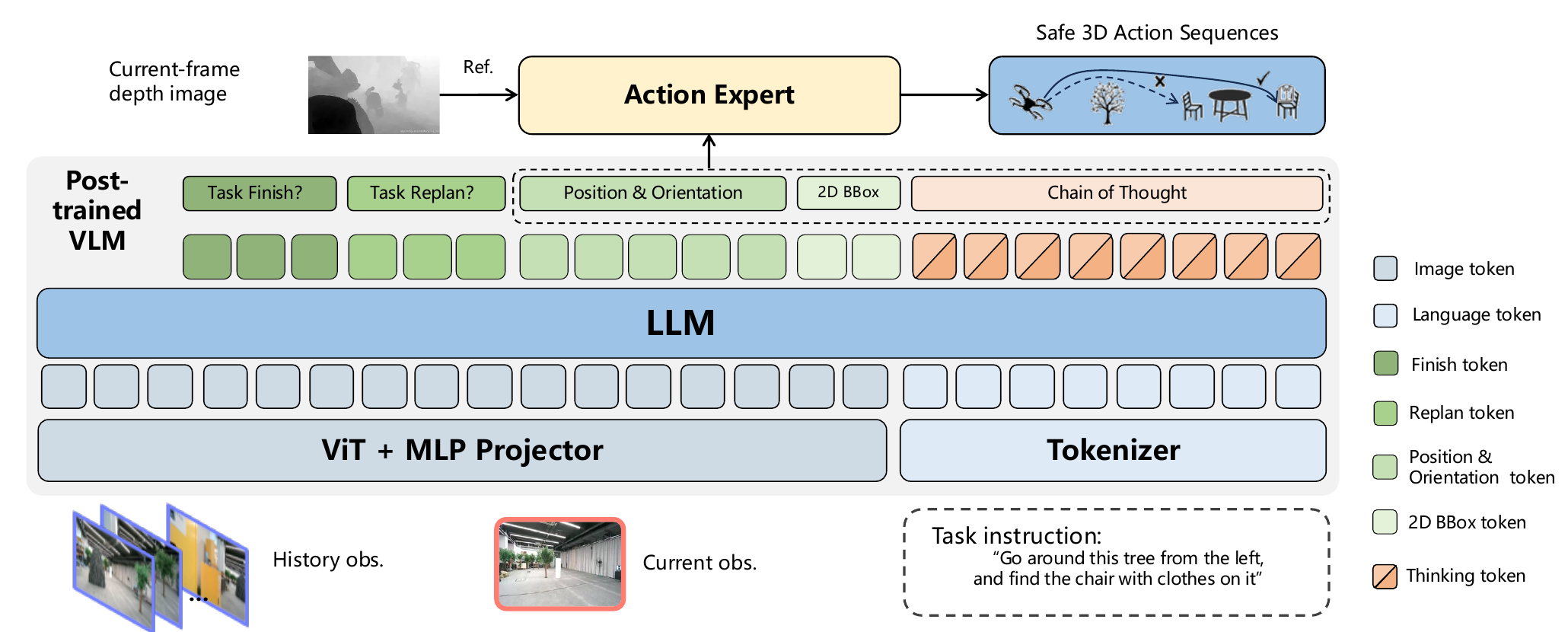}
      \caption{The model architecture of the proposed VLN-AR model.}
      \label{fig:model}
      \vspace{-0.3cm}
    \end{figure*}
    
    \vspace{-0.1cm}
    \subsection{Onboard Deployment for Robotics}
    Driven by advances in edge computing, onboard deployment of foundation models on resource-constrained robotic platforms, particularly UAVs, has attracted increasing attention. Common approaches include quantization, pruning, knowledge distillation, and operator fusion to reduce latency and meet \add{deployment requirements} \cite{ma2025runningvlasrealtimespeed}. However, preserving \add{multimodal capabilities under strict computational and energy constraints remains challenging for UAVs}.
    
    Existing robotic foundation models such as OpenVLA \cite{kim2024openvlaopensourcevisionlanguageactionmodel} and the $\pi_0$ family \cite{black2024pi0visionlanguageactionflowmodel} typically rely on high-performance GPUs and are difficult to deploy onboard UAVs. Although lightweight architectures, policy distillation \cite{rusu2016policydistillation, sink}, mixed-precision inference \cite{micikevicius2018mixedprecisiontraining}, and hardware-aware optimization have been explored, maintaining stable closed-loop control under aggressive compression remains an open problem. This highlights the need for \add{efficient UAV-oriented models and onboard inference frameworks}.

\section{Model and Data}
    \label{sec:Model_and_Data}
    Intelligent aerial navigation requires UAVs to autonomously operate in unknown environments by interpreting open-vocabulary language instructions and onboard visual observations.
    To this end, we propose VLN-AR, a hierarchical vision-language-navigation framework in which a unified VLM $\pi_{VLM}$ generates a token-level policy and a learning-free action expert $\phi_{\text{LF}}$ performs safe low-level flight commands (Fig.~\ref{fig:model}). The overall system is defined as:
    \begin{equation}
    \pi_{\text{VLN-AR}} = \phi_{\text{LF}} \circ \pi_{VLM}.
    \end{equation}
    Formally, given a language instruction $L$ and multimodal observations $\mathcal{O}_{t-k:t}$, our VLN-AR framework maps vision-language inputs to flight actions $a_t$:
    \begin{equation}
        \begin{aligned}
        a_t = \pi_{\text{VLN-AR}}(L, \mathcal{O}_{t-k:t}).
        \end{aligned}
    \end{equation}

    \begin{figure*}[ht]
      \centering
      \includegraphics[width=0.82\linewidth]{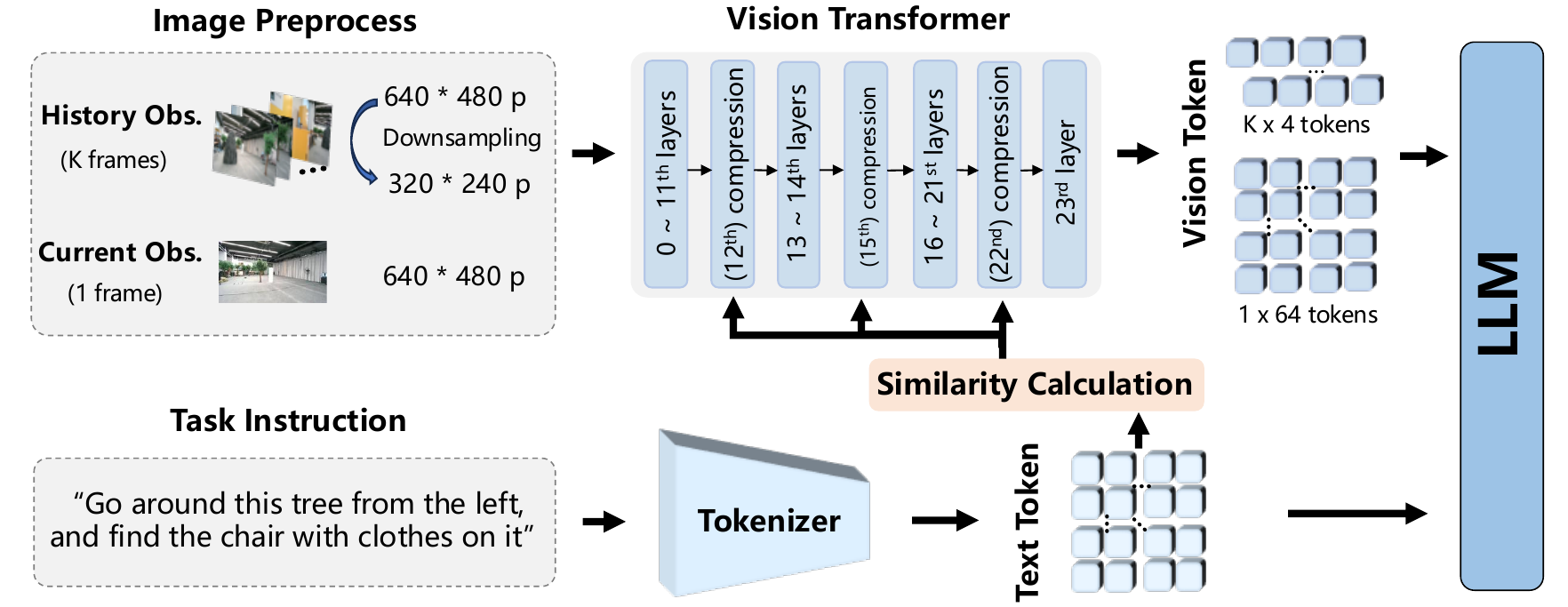}
      \caption{The improved ViT architecture and token \add{sparsification}.} 
      \label{fig:vit}
      \vspace{-0.3cm}
    \end{figure*}

    \subsection{Model Architecture}
    
    \subsubsection{Hierarchical Architecture}
    \label{sec:model}
    
    VLN-AR adopts a hierarchical architecture that combines a post-trained VLM with an efficient action expert, enabling tight coupling between high-level semantic reasoning and low-level execution safety, as shown in Fig.~\ref{fig:model}.
    Formally, given multimodal observations $\mathcal{O}_{t-k:t} = \{I_{t-k:t-1}^{rgb}, I_t^{rgb}, p_t\}$, where $I_{t-k:t-1}^{rgb}$ denotes the historical RGB observations from time $t-k$ to $t-1$, $I^{rgb}_t$ denotes the current RGB observation at time $t$.
    The objective is to learn a VLN model $\pi_{\text{VLN-AR}}$ that generates an action sequence $a_{t}$ maximizing the task success probability $P(\texttt{success}|L, \mathcal{O}_{t-k:t})$:
    \begin{equation}
        \begin{aligned}
        & \pi_{\text{VLN-AR}} = \\ 
        &\arg\max_{\pi} \; \mathbb{E}_{a_{t} \sim \pi(\cdot \mid L, \mathcal{O}_{t-k:t})}
        \left[ P\!\left(\texttt{success} \mid L, \mathcal{O}_{t-k:t}, a_{t}\right) \right].
        \end{aligned}
    \end{equation}
    The proposed VLM $\pi_{\text{VLM}}$ performs multimodal reasoning and navigation prediction $y_t$ as
    \begin{align}
        y_t = [ {y}^{\text{pose}}_{\text{t}}, {y}^{\text{bbox}}_{\text{t}}, {y}^{\text{reason}}_{\text{t}},{y}^{\text{finish}}_{\text{t}}, {y}^{\text{replan}}_{\text{t}} ] = \pi_{\text{VLM}}(L, \mathcal{O}_{t-k:t}),
    \end{align}
    where ${y}^{\text{pose}}_{\text{t}}$ denotes the predicted 3D navigation waypoints with the desired yaw orientations in the body coordinate frame, ${y}^{\text{bbox}}_{\text{t}}$ represents the predicted 2D bounding box in the image plane, ${y}^{\text{reason}}_{\text{t}}$ represents the CoT reasoning-based navigation intent, ${y}^{\text{replan}}_{\text{t}}$ indicates whether the referenced task is replanned in the current observation, and ${y}^{\text{finish}}_{\text{t}}$ represents whether the referenced task is completed.     
    By tightly coupling semantic reasoning with fine-grained spatial understanding, $\pi_{\text{VLM}}$ enables open-vocabulary and long-horizon navigation. 
    
    To ensure safe and reliable flight execution, $\pi_{\text{VLN-AR}}$ integrates a learning-free action expert $\phi_{\text{LF}}$, enhanced based on Ego-Planner \cite{9309347}, as illustrated in Fig.~\ref{fig:model}:
    \begin{align}
        a_t = \phi_{\text{LF}}({y}^{\text{pose}}_{\text{t}}, {y}^{\text{bbox}}_{\text{t}}, {y}^{\text{reason}}_{\text{t}}).
        \label{eq:ae}
    \end{align}
    Given the navigation target location, this module validates and refines the locally generated trajectory, thereby improving obstacle avoidance and reducing execution time in cluttered environments. Compared with large generative action experts (e.g., the 0.3B flow-matching action expert in the $\pi_0$ \cite{black2024pi0visionlanguageactionflowmodel}), the proposed module runs at a control frequency exceeding 200 Hz on an edge computation device, enabling reliable high-speed flight and low-latency collision-free navigation in complex and previously unseen environments.
    As formulated in Eq.~(\ref{eq:ae}), the action expert supports three types of instruction inputs.    
    \add{The model adaptively selects the target representation according to the task type. First, for some navigation tasks such as object navigation, it preferentially uses a 2D bounding box to localize the target and combines it with the average depth within the detected region to predict a 3D waypoint and yaw angle in the body frame for the action expert. Second, for spatial-relation tasks, such as moving to the right side of an object or passing through the circle, the desired position may lie outside the visible target region and cannot be reliably inferred from the bounding box and depth map. The VLM therefore directly predicts 3D target poses, including waypoints and yaw angles, in the body frame, with position measured in meters and yaw in degrees. Third, when the $360^\circ$ observation lacks sufficient task-relevant information, the post-trained VLM \addagain{performs CoT reasoning} over accumulated observations to identify a promising candidate region and \addagain{directly} generates the corresponding waypoint and heading for exploration.}

    \add{In summary, the 3D pose, 2D BBox, and CoT outputs are used to generate or refine the 3D poses and yaw information sent to the action expert. In contrast, the ${y}^{\text{replan}}_{\text{t}}$ and ${y}^{\text{finish}}_{\text{t}}$ flags are used only for high-level task-state assessment and process scheduling and are not directly passed to the low-level action-execution module.}

    \begin{figure*}[ht]
      \centering
      \includegraphics[width=0.82\linewidth]{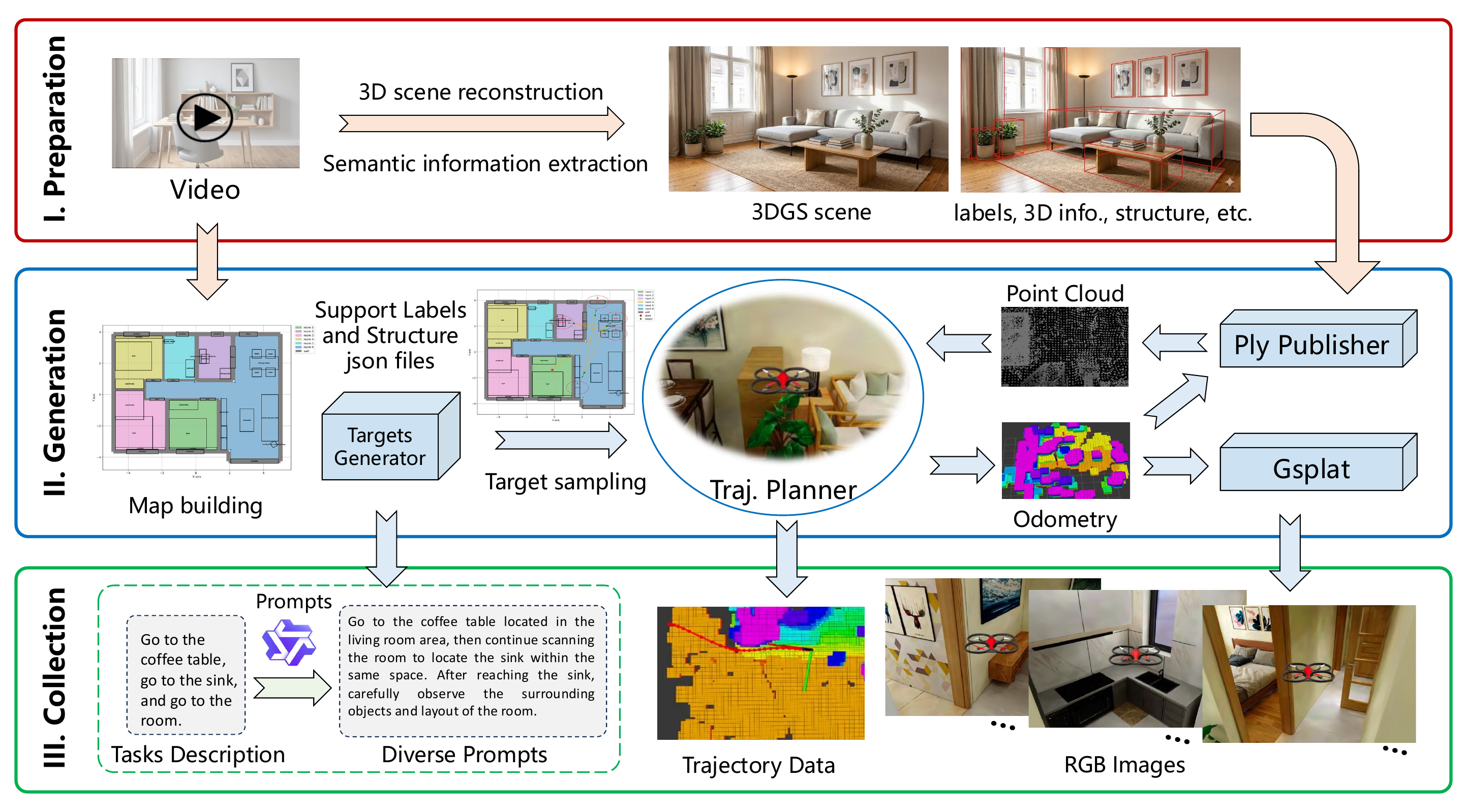}
      \caption{Automated UAV navigation data generation and collection process.}
      \label{fig:data_pipeline}
      \vspace{-0.3cm}
    \end{figure*}

    \subsubsection{Instruction-guided Token Sparsification in Vision Transformer}
    \label{sec:vit}
    
    To improve efficiency under multi-image inputs and enable real-time deployment on resource-constrained UAV platforms, we introduce an instruction-guided token sparsification mechanism within the Vision Transformer (ViT) of $\pi_{\text{VLM}}$, as illustrated in Fig.~\ref{fig:vit}. Empirically, we observe that a significant portion of inference latency is concentrated in the ViT stage on lightweight embedded computing platforms (e.g., NVIDIA Jetson Orin NX), primarily due to the quadratic complexity of self-attention over dense token sets. Moreover, real-world navigation images often contain large task-irrelevant regions (e.g., sky, repetitive textures), leading to unnecessary computation and degraded attention efficiency. Therefore, our goal is to jointly reduce redundant visual computation while preserving navigation-critical spatial semantics. \add{In this work, the interval between historical frames is 0.5 s, the number of historical frames is $K=8$, and the total duration covered by the historical frames is 4 s.}
    
    Our instruction-guided progressive token sparsification strategy is inspired by recent layer-wise visual token compression studies \cite{li2025qgvtcquestionguidedvisualtoken,cai2025flashvlmtextguidedvisualtoken,liu2025lacoefficientlayerwisecompression}. 
    From a system perspective, the input consists of multi-frame historical observations and the current RGB frame. Historical observations are first temporally aggregated and downsampled (e.g., from $640 \times 480$p to $320 \times 240$p) to reduce redundancy, while the current observation is kept at full resolution. Both are processed through a hierarchical ViT backbone with periodic feature compression layers, yielding a structured set of vision tokens.    
    Importantly, the sparsification is guided by task semantics rather than purely visual heuristics, ensuring that navigation-critical regions are preserved. 
    \add{Specifically, the task instruction is represented using the pretrained Qwen3-VL language embeddings, while the intermediate visual features are projected into the same embedding space using the pretrained Qwen3-VL vision projection modules. Let $\bar{\mathbf{v}}^{(l)}_i$ and $\bar{\mathbf{t}}_j$ denote the $L_2$-normalized projected visual and text embeddings, respectively. Their cosine similarity is computed as $s^{(l)}_{i,j}=(\bar{\mathbf{v}}^{(l)}_i)^\top\bar{\mathbf{t}}_j$, and the visual-token relevance is obtained by max pooling, $r^{(l)}_i=\max_j s^{(l)}_{i,j}$.}
    
    Recent studies \cite{liu2025lacoefficientlayerwisecompression,li2025qgvtcquestionguidedvisualtoken} have shown that intermediate-to-deep ViT layers are more suitable for token compression than shallow layers, and that progressive in-encoder compression achieves a better accuracy--efficiency trade-off. Motivated by this insight, we introduce compression modules at \add{ViT blocks indexed 12, 15, and 22 using zero-based indexing} to progressively reduce redundant visual tokens while preserving navigation-critical spatial semantics. The resulting vision tokens are then passed to the LLM for downstream reasoning.
    
    \add{At each compression stage, visual tokens are ranked according to $r^{(l)}_i$ and selected using a Top-$K$ strategy with predefined stage-wise retention ratios rather than an absolute similarity threshold. The selected tokens retain their original spatial indices during subsequent processing. The cumulative token-retention ratios after the three stages are set to $75\%$, $50\%$, and $25\%$ of the original token count, respectively.}
    This design enables early semantic filtering within the ViT, rather than relying on full encoding followed by pruning, thereby significantly reducing redundant computation. 
    
    \add{Overall, the final $25\%$ retention reduces vision tokens by $75\%$ while maintaining navigation performance, as validated in Fig.~\ref{fig:vit_analysis}.}
    \add{The proposed token sparsification is enabled only during model inference and remains disabled during SFT and RFT.}

    \subsection{Automated High-Fidelity Data Collection}
    \label{sec:3DGS}
    
    High-fidelity 3D trajectory data for complex UAV navigation remains relatively scarce, primarily due to challenges such as the low level of automation in data collection \cite{liu2023aerialvlnvisionandlanguagenavigationuavs}, the high cost of real-world data acquisition \cite{wang2025uavflowcolosseorealworldbenchmark}, and the limited realism of simulated data. To address these issues, we first develop an efficient simulation environment based on 3DGS method and the Unity Engine. The proposed simulator is capable of generating and reconstructing indoor environments with near-photorealistic visual quality, particularly in terms of lighting conditions, texture fidelity, and fine-grained visual features. 
    Furthermore, we augment the training corpus with a small amount of additional real-world UAV data to improve robustness and sim-to-real generalization.


    The automated pipeline incorporates UAV dynamics and comprises scene reconstruction and annotation, automated navigation generation, and multimodal data collection (Fig.~\ref{fig:data_pipeline}). It produces photorealistic observations and dynamically feasible trajectories and can be launched with a single script.
    
    \subsubsection{Stage I. Scene Reconstruction and Annotation}    
    Unlike traditional mesh-based rendering approaches, we employ 3DGS to capture real-world illumination variations and continuous geometric details. We first collect video streams covering diverse indoor (e.g., offices and corridors) and outdoor (e.g., parks and streets) environments, which are then converted into high-quality 3DGS scene representations and imported into the Unity Engine.     
    In addition to 3DGS scene reconstruction, we leverage scene understanding tools (e.g., InstaScene~\cite{yang2025instascenecomplete3dinstance}) to extract various scene attributes, including structural layout information, object labels, corresponding 3D bounding boxes, scene maps, and more. These semantic annotations serve as core inputs for subsequent automated task generation and data acquisition.
    
    \subsubsection{Stage II. Automated Navigation Generation}    
    Building upon the rich semantic annotations described above, we develop a closed-loop automated data collection pipeline. The detailed procedure is described below.
    
    {(a) Task definition and waypoint randomization:}    
    Instead of directly generating task instructions, we adopt a reverse-engineering strategy to generate navigation tasks. Given the scene map, collision-free and safe regions are first identified, from which start and goal positions are randomly sampled. The corresponding orientations are then generated accordingly to ensure kinematic feasibility. 
    After defining valid navigation tasks, language instructions (e.g., ``Go to the coffee table, go to the sink, and go to the room'') are automatically synthesized using a set of rule-based templates or VLMs. By constraining the sampling process within expert-defined feasible regions, we ensure that the generated tasks are both sufficiently challenging and reliably solvable.
    
    {(b) Dynamics-constrained motion planning:}    
    In simulation, a UAV equipped with an RGB, depth, and lidar sensor employs a numerical optimization-based trajectory planner \cite{9765821} that explicitly accounts for obstacle avoidance, height constraint, dynamical feasibility, trajectory smoothness, and time optimality to generate flight trajectories. The planner fully exploits the UAV's maneuverability, enabling not only horizontal obstacle avoidance but also vertical motion along the $Z$-axis to bypass obstacles, thereby achieving collision-free navigation in complex 3D environments.
    
    \subsubsection{Stage III. Multimodal Navigation Data Collection} Once a feasible trajectory has been generated, the system executes the flight in simulation and records synchronized multimodal observations. During this process, the UAV tracks the planned trajectory using an adaptive flight controller~\cite{wu2026hand}, while continuously collecting onboard visual inputs and corresponding state measurements (e.g., position, velocity, and attitude commands). 
    
    {(a) Synchronized data collection:} 
    Within the Unity simulation, the UAV receives odometry information published by the SE(3) simulator to update its pose in real time. During autonomous flight along the planned trajectory, the system records multimodal data at 10 Hz, including ego-centric RGB images, depth images, lidar point clouds, and the UAV's 6-DoF odometry in the world coordinate frame. In addition, data augmentation techniques such as trajectory randomization and observation randomization are applied to enhance data diversity and prevent overfitting during training.
    
    {(b) Instruction diversity enhancement:}    
    After collecting RGB images and flight trajectories, we further refine task instructions using multi-frame visual context and an MLLM \add{such as Qwen3-VL-32B-Instruct \cite{bai2025qwen3}} to enhance the diversity of language annotations. In this step, a video understanding model is employed to re-describe the collected visual data in a more diverse and semantically enriched form. For example, given the collected trajectory data and corresponding first-person view images, the instruction ``Go to the coffee table, go to the sink, and go to the room'' can be reasoned and refined as ``Go to the coffee table located in the living room area, then continue scanning the room to locate the sink within the same space. After reaching the sink, carefully observe the surrounding objects and layout of the room.''

    \add{{(c) Filtering and validation:}    
    For trajectory-level validation, we compare the trajectory endpoint with the ground-truth waypoint from simulation. Samples with errors below 1 m are considered successful; otherwise, they are discarded. After prompt rewriting, Doubao 1.8 Pro \cite{bytedance2025seed18} automatically verifies consistency in the target, trajectory, spatial relation, and task constraints.}

\section{Training and Deployment}
    \label{sec:Methods}
    
    \subsection{Multi-stage Training Recipe}
    
    As described in Sec.~\ref{sec:model}, we formulate aerial navigation as a unified multimodal policy learning problem, where a VLM parameterizes a token-level navigation policy $\pi_{\text{VLM}}$:
    \begin{equation}
        \pi_{\text{VLM}}({y}_t \mid {L}, \mathcal{O}_{t-k:t}).
    \end{equation}
    Based on the multi-source data, we design a progressive three-stage training recipe that systematically improves cross-task generalization for aerial navigation, as illustrated in Fig.~\ref{fig:overview}(II). The baseline model adopts the Qwen multimodal model series, such as Qwen3-VL-4B. The proposed framework progressively enhances the model’s spatial understanding, task planning, and high-level reasoning capabilities in complex aerial environments.
    
    \subsubsection{Stage I. Grounding-reasoning-enhanced SFT}

    In the first stage, we perform full-parameter SFT on a mixture of large-scale datasets, including VQA, object grounding, spatial reasoning, and reasoning tasks. The objective is to learn a unified multimodal policy capable of outputting structured grounding and reasoning tokens, as illustrated in Fig.~\ref{fig:overview}.
    The Stage I training objective loss function is defined as:
    \begin{equation}
        \begin{aligned}
        \mathcal{L}_{\text{stageI}} =
        & \lambda_{\text{vqa}} \mathcal{L}_{\text{CE}}(\mathbf{y}^{\text{vqa}}_{\text{pred}}, \mathbf{y}^{\text{vqa}}_{\text{gt}}) \\
        +& \lambda_{\text{bbox}} \mathcal{L}_{\text{CE}}(\mathbf{y}^{\text{bbox}}_{\text{pred}}, \mathbf{y}^{\text{bbox}}_{\text{gt}}) \\
        +& \lambda_{\text{reason}} \mathcal{L}_{\text{CE}}(\mathbf{y}^{\text{reason}}_{\text{pred}}, \mathbf{y}^{\text{reason}}_{\text{gt}}),
        \end{aligned}
    \end{equation}
    where $\mathcal{L}_{\text{CE}}$ denotes the cross-entropy loss for discrete token prediction, and $\mathbf{y}^{\text{vqa}}_{\text{pred}}$, $\mathbf{y}^{\text{vqa}}_{\text{gt}}$ represent the predicted and ground-truth tokens for VQA data outputs, respectively. Similarly, $\mathbf{y}^{\text{bbox}}_{\text{pred}}$ and $\mathbf{y}^{\text{bbox}}_{\text{gt}}$ denote the predicted and ground-truth grounding tokens, $\mathbf{y}^{\text{reason}}_{\text{pred}}$ and $\mathbf{y}^{\text{reason}}_{\text{gt}}$ denote the predicted and ground-truth CoT reasoning tokens. And $\lambda_{\text{vqa}}$, $\lambda_{\text{bbox}}$, and $\lambda_{\text{reason}}$ are weighting coefficients.

    \subsubsection{Stage II. Navigation-specific SFT}

    In the second stage, we further fine-tune the model using high-quality UAV navigation trajectories, mixed with spatial reasoning data. This stage focuses on improving robustness in complex aerial navigation, including pose prediction, task completion, replanning behavior, etc.
    The optimization objective loss function is:
    \begin{equation}
        \begin{aligned}
        \mathcal{L}_{\text{stageII}} =
        & \lambda_{\text{pose}} \mathcal{L}_{\text{reg}}(\mathbf{y}^{\text{pose}}_{\text{pred}}, \mathbf{y}^{\text{pose}}_{\text{gt}}) \\
        +& \lambda_{\text{finish}} \mathcal{L}_{\text{CE}}(y^{\text{finish}}_{\text{pred}}, y^{\text{finish}}_{\text{gt}}) \\
        +& \lambda_{\text{replan}} \mathcal{L}_{\text{CE}}(y^{\text{replan}}_{\text{pred}}, y^{\text{replan}}_{\text{gt}}).
        \end{aligned}
    \end{equation}
    where $\mathcal{L}_{\text{reg}}$ denotes a regression loss (e.g., L1 loss or Smooth L1 loss), and $\lambda_{\text{pose}}$, $\lambda_{\text{finish}}$, and $\lambda_{\text{replan}}$ are weighting coefficients.

    \subsubsection{Stage III. Reinforcement Fine-tuning for Reasoning-aware Navigation}

    Deploying a VLN system requires tight coupling between high-level multimodal reasoning and low-level control execution. In this stage, we integrate the VLM and the action expert into a closed-loop simulation environment for reinforcement fine-tuning (RFT), as illustrated in Fig.~\ref{fig:overview}(II). Specifically, we adopt an online RFT framework based on Group Relative Policy Optimization (GRPO)\cite{shao2024deepseekmathpushinglimitsmathematical} to improve navigation accuracy, robustness, and decision consistency in complex 3D environments.
    
    For each navigation episode, the model generates a navigation command sequence conditioned on the language instruction and historical observations.  
    We denote the generated trajectory as
    \begin{equation}
        \mathbf{\tau} \sim \pi_\text{VLN-AR}(\mathbf{\tau} \mid L, \mathcal{O}_{t-k:t}).
    \end{equation}
    We further employ a sparse outcome-level reward defined as
    \begin{equation}
    R(\mathbf{\tau}) =
        \begin{cases}
        1, & \text{if } \texttt{success}(\mathbf{\tau}), \\
        0, & \text{otherwise},
        \end{cases}
    \end{equation}
    where $\texttt{success}(\mathbf{\tau})$ indicates successful task completion. This sparse binary reward avoids heuristic reward shaping and directly optimizes navigation success.
    
    Following GRPO, for each instruction we sample a trajectory group $\mathcal{G} = \{\mathbf{\tau}_1,\dots,\mathbf{\tau}_G\}$ and compute the normalized group-relative advantage:
    \begin{equation}
        \hat{A}(\mathbf{\tau}_i,\mathcal{G}) =
        \frac{
        R(\mathbf{\tau}_i)-\mu_{\mathcal{G}}
        }{
        \sigma_{\mathcal{G}}+\epsilon
        },
    \end{equation}
    where $\mu_{\mathcal{G}}$ and $\sigma_{\mathcal{G}}$ denote the mean and standard deviation of rewards within the group, and $\epsilon$ is a small numerical stabilizer (e.g., $1e{-}8$).
    For simplicity, we use $\pi_{\theta}$ to denote $\pi_{\text{VLN-AR}}$. The GRPO objective is formulated as:
    \begin{equation}
    \begin{aligned}
    \mathcal{L}_3 = &
    - \mathbb{E}_{\tau \sim \pi_{\theta_{\text{old}}}}
    \Bigg[
    \sum_{i=1}^{G}
    \min \Big(
    r_i(\theta)\hat{A}(\tau), \\
    &\mathrm{clip}\!\left(
    r_i(\theta), 1-\epsilon, 1+\epsilon
    \right)\hat{A}(\tau)
    \Big)
    \Bigg]
    + \beta \, \mathrm{KL}
    \!\left(
    \pi_{\theta} \,\|\, \pi_{\text{ref}}
    \right),
    \end{aligned}
    \end{equation}
    where
    \begin{equation}
        r_i(\theta)=
        \frac{
        \pi_{\theta}(\mathbf{\tau}_i \mid L,\mathcal{O}_{t-k:t})
        }{
        \pi_{\theta_{\text{old}}}(\mathbf{\tau}_i \mid L,\mathcal{O}_{t-k:t})
        }.
    \end{equation}
    
    Here, $\pi_{\theta_{\text{old}}}$ denotes the old policy used for trajectory sampling, $\pi_{\mathrm{ref}}$ is a frozen reference policy for KL regularization, $\varepsilon$ is the clipping threshold, and $\beta$ controls the KL penalty strength.
    Since sparse binary rewards frequently lead to zero-variance groups ($\sigma_{\mathcal{G}} = 0$) when all sampled trajectories either succeed or fail, the group-relative advantage becomes uninformative. Following common practice in group-based RL methods \cite{shao2024deepseekmathpushinglimitsmathematical}, we discard such degenerate groups during training to avoid gradient collapse. 

    To better support long-horizon reasoning-aware navigation tasks, we additionally introduce a stage-wise reward accumulation mechanism. Specifically, for a trajectory $\tau$ consisting of $N$ sequential reasoning-execution stages, the final reward is defined as:
    \begin{equation}
        R(\tau)=\sum_{j=1}^{N}\gamma^{j-1}R_j(\tau),
    \end{equation}
    where $R_j(\tau)$ denotes the sparse success reward at the $j$-th stage, and $\gamma \in (0,1]$ is a discount factor for long-horizon credit assignment. The accumulated reward is subsequently used in the group-relative advantage computation, enabling more stable optimization for long-horizon navigation tasks with delayed outcomes.


    \begin{figure*}[htp]
      \centering
      \includegraphics[width=0.82\linewidth]{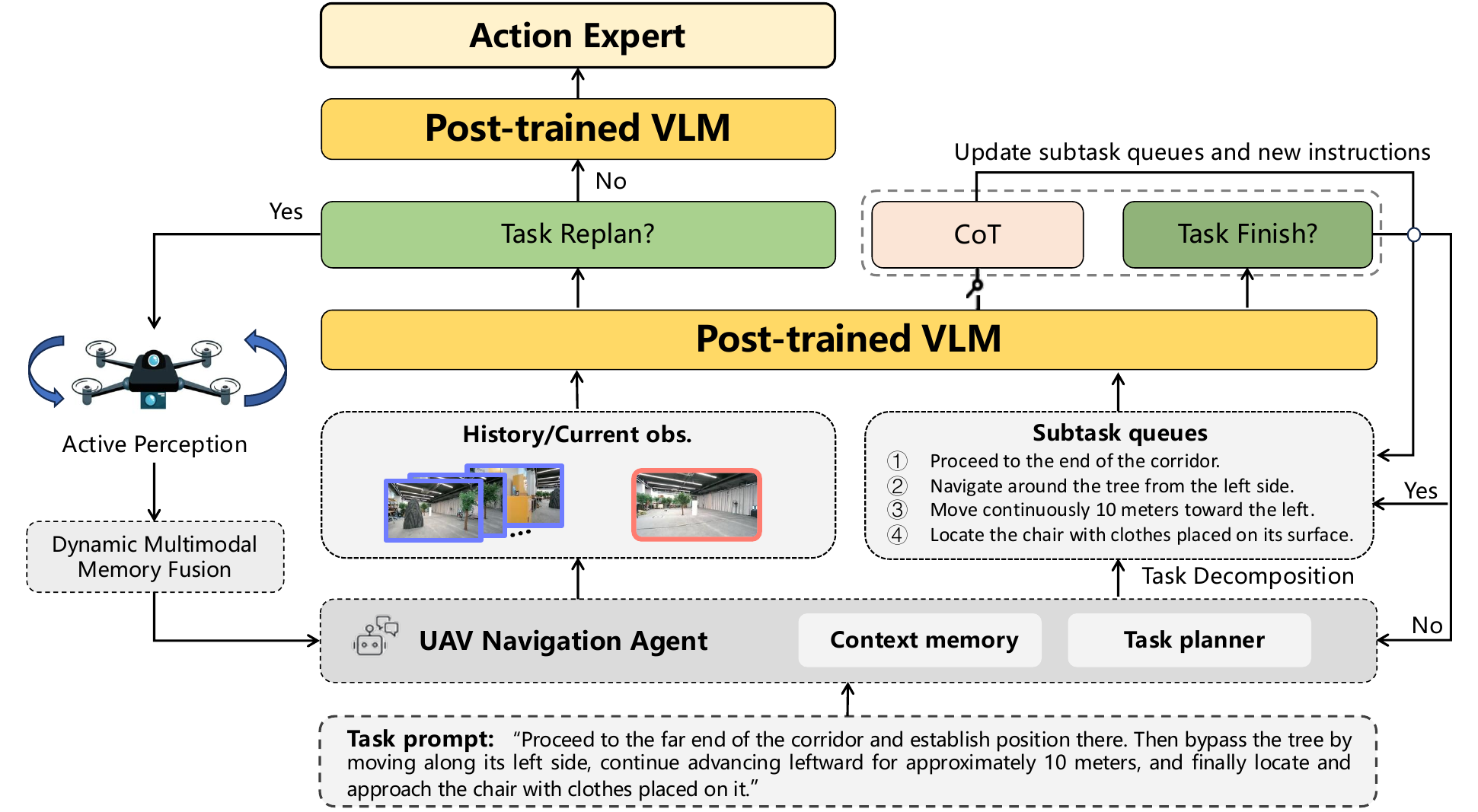}
      \caption{Cognitive-driven UAV navigation agent pipeline for VLN-AR.}
      \label{fig:agent}
      \vspace{-0.2cm}
    \end{figure*}

    \begin{algorithm}[!t]
    \caption{\textbf{Cognitive-Driven UAV Navigation Agent}}
    \vspace{-2mm}
    \hrule
    \label{alg:vla_agent}
    \begin{algorithmic}[1]
    
    \Require Navigation instruction $L$, Historical observations $\mathcal{H}$, Current observation $\mathcal{O}_t$, Context memory $\mathcal{C}$, Reasoning switch $r \in \{\text{think\_on}, \text{think\_off}\}$
    \Ensure Executable actions and mission status
    
    \State Initialize dynamic multimodal memory $\mathcal{M}$
    \State Initialize subtask queue $\mathcal{Q} \leftarrow \emptyset$
    \State Initialize context memory $\mathcal{C} \leftarrow \emptyset$
    \State Subtask queues $s$ $\leftarrow$ TaskPlanner$(L)$
    \State Mission status $\leftarrow$ incomplete
    
    \While{mission not completed}
    
        \State Capture current observation $\mathcal{O}_t$
        \State Update memory: $\mathcal{H}, \mathcal{M} \leftarrow  \text{Perception}(\mathcal{O}_t, \mathcal{H}, \mathcal{M})$
    
        \State $\text{replan}, \text{finish}, \text{cot} \leftarrow \text{PostTrainedVLM}(L, \mathcal{O}_{t-k:t}, r)$
    
        \If{all subtasks completed}
            \State Mission $\leftarrow$ completed \
            \State \textbf{Break}
        \EndIf
    
        \If{subtask $s$ completed ($\text{finish} = \text{True}$)}
            \State Pop($\mathcal{Q}$)
            \State \textbf{Continue}
        \EndIf
        
        \If{$\text{replan} = \text{False}$}
            \State $a_t \leftarrow \text{VLN-AR}(L, \mathcal{C})$ \add{(asynchronous, 2 Hz)}
        \Else
            \State Perform active perception (view exploration)
            \State $w \leftarrow \text{ExplorationPolicy}(\mathcal{C})$
            \State $\psi \leftarrow \text{DefaultYaw}$
            \State $\mathcal{C} \leftarrow \text{DynamicMultimodalMemoryFusion}(\mathcal{H}, \mathcal{M})$
            \If{target still not detected}
                \State Invoke reasoning module (CoT)
                \State $a_t \leftarrow \text{VLN-AR}_{CoT}(L, \mathcal{C})$ \add{(asynchronous, 2 Hz)}
            \EndIf
        \EndIf

        \State Execute and replan trajectory $a_t$ \add{(asynchronous, $\textgreater$200 Hz)}

        \If{$r = \text{think\_on}$}
            \State Update subtask queues via CoT:
            \State $\mathcal{Q} \leftarrow \text{UpdateQueue}(\mathcal{Q}, \text{cot})$
            \State $s \leftarrow \text{Head}(\mathcal{Q})$
        \EndIf
    
    \EndWhile
    
    \end{algorithmic}
    \end{algorithm}

    \subsection{Cognitive-Driven Aerial Navigation Agent}

    To enable UAVs to achieve efficient decision-making in complex long-horizon tasks, we integrate a cognitive-driven aerial navigation agent into the VLN-AR framework, as shown in Fig.~\ref{fig:agent} \add{and Alg.~\ref{alg:vla_agent}. The proposed agent operates at 2 Hz, corresponding to a VLM inference rate of 2 Hz, whereas the action expert runs at over 200 Hz. VLM inference and action execution are performed asynchronously.} This agent can incorporate efficient heuristic rules and a dynamic memory fusion module to compensate for intelligent behaviors that are insufficiently captured by the VLN model, thereby enabling task management and system-level autonomy. Specifically, this UAV agent features the following three core mechanisms.

    \subsubsection{Autonomous Decomposition for Long-horizon Tasks}    
    When faced with complex natural language instructions containing multiple sub-goals, the agent demonstrates exceptional autonomous task planning capabilities. Leveraging the robust logical comprehension of large models, it automatically decomposes raw instructions into a series of temporally coherent sub-tasks, as illustrated in Fig.~\ref{fig:agent}. During flight, the agent autonomously evaluates the completion status of the current sub-task based on real-time visual feedback and, upon completion, seamlessly triggers the transition to the next sub-task, ensuring the closed-loop execution of long-horizon navigation, as shown in Fig.~\ref{fig:agent}.

    \subsubsection{Dynamic Multimodal Memory Fusion}    
    The agent maintains a dynamic memory buffer, autonomously and continuously recording and integrating historical task instructions, past flight trajectories, key visual frames, and task states. By aggregating spatio-temporal context information across time, the agent establishes a persistent cognition of the dynamic perception, which effectively prevents state disorientation and maintains the consistency of navigation intentions when encountering temporary visual occlusions or visually similar scenes. Consequently, the system can handle memory-dependent tasks, such as “navigate to the previously visited white dining table” or "return to the initial starting point."
    
    \subsubsection{Heuristic Exploration in Highly Challenging Scenarios}
    In highly unstructured environments, the target object may lie outside the initial field of view or the unexplored region. To address this, the agent is equipped with a condition-triggered reasoning capability. When standard search strategies fail (e.g., a full $360^\circ$ in-place rotation for active perception still yields $y^\text{replan}_{t} = \text{true}$), the agent temporarily suspends regular waypoint generation. Instead, based on the currently observed environmental context (e.g., surrounding image frames), historical memory, and commonsense priors, it leverages the post-trained VLM's CoT reasoning to infer the most probable hidden regions where the target may reside, as shown in Fig.~\ref{fig:agent}. \add{For example, the CoT reasoning module may infer that “the corridor on the left is more likely to lead to the target gate in our task.” This reasoning result is subsequently converted into the corresponding 3D waypoints and yaw angles, which are passed to the action expert for execution.} Guided by this reasoning process, the agent generates exploratory waypoints to actively search the unknown environment.
    \add{If no plausible exploration direction can be identified, the CoT module marks the current task as unreachable and updates the high-level task sequence accordingly, for example by proceeding to the next task.}

    \add{Furthermore, Fig.~\ref{fig:agent} and Alg.~\ref{alg:vla_agent} illustrate the workflow of the proposed agent. The key components are the assessment and feedback mechanisms for task completion ($y^\text{finish}_{t}$) and task replanning ($y^\text{replan}_{t}$). Specifically, the $y^\text{finish}_{t}$ flag indicates task completion. After each new observation, the system evaluates the current subtask: if \addagain{the VLM predicts} $y^\text{finish}_{t}=true$ \addagain{at the current decision step}, it terminates the subtask and proceeds to the next; otherwise, execution continues.
    The system then assesses whether the observation provides sufficient task-relevant information. If so, $y^\text{replan}_{t}=false$, and the model selects an appropriate target representation based on the task type, passing the resulting pose and yaw to the action expert.
    If the observation is insufficient, $y^\text{replan}_{t}=true$, and the UAV actively explores the surroundings through a continuous $360^\circ$ rotation. Once the target is detected, $y^\text{replan}_{t}$ becomes false and the corresponding command is generated. If no target is found after a full rotation, CoT reasoning identifies a promising exploration region and outputs a waypoint and heading for execution. If no plausible direction exists, the subtask is marked unreachable, and the system proceeds to the next one.}

    \subsection{System-Level Onboard Deployment Optimization}
    \label{sec:onboard}
    To enable real-world deployment on general UAV platforms, we design a lightweight inference framework tailored for severely resource-constrained airborne computing systems. Due to strict limitations in payload, power consumption, and flight endurance, high-performance general-purpose computing platforms are difficult to deploy directly. 
    In this section, we adopt the NVIDIA Jetson Orin NX platform\footnote{\url{https://www.nvidia.com/en-us/autonomous-machines/embedded-systems/jetson-orin/}} as the onboard computing platform and perform system-level co-optimization under its constraints in computation throughput and memory bandwidth. After structural integration and payload pruning, the total weight of the perception–computation module is reduced to approximately 80g, making it suitable for micro-UAV deployment. However, executing a full foundation model on such a platform still faces a fundamental trade-off between computational capacity and real-time performance.
    To address this issue, we design an onboard inference acceleration strategy from two complementary perspectives: reducing computational redundancy and improving system execution efficiency, consisting of the following three mechanisms.
    
    First, at the algorithmic level, we reduce computational redundancy in the ViT by introducing visual token sparsification, as described in Sec.~\ref{sec:vit}. \add{Importantly, the visual token sparsification is introduced during deployment optimization, not during training.} This reduces the input token length at the source, thereby alleviating downstream inference cost and mitigating the computational complexity inherent in Transformer-based architectures.
    Second, at the system execution level, we optimize the ViT inference pipeline to address limited operator efficiency under the ARM architecture. Specifically, we perform structural optimization including attention computation reordering, feature dimension restructuring, and operator scheduling refinement to improve parallelism and cache utilization. In addition, we accelerate key matrix operations via SIMD instructions and optimized memory access patterns, improving computational efficiency under constrained power budgets.
    On top of this, we introduce a set of system-level inference graph optimizations, including Flash-Attention for efficient attention computation, fusion of Feed-Forward Networks (FFN) with normalization layers to reduce intermediate memory traffic, and KV-cache preloading to improve memory bandwidth utilization. Furthermore, CUDA Graphs are employed to statically schedule computation graphs, reducing kernel launch overhead and improving pipeline stability.
    
    Finally, we incorporate a heterogeneous low-precision quantization strategy along the model execution path. Both model weights and dynamic activations (including KV-cache) are uniformly quantized, significantly reducing memory footprint and improving inference throughput, while preserving spatial grounding and localization accuracy as much as possible.

\section{Results}
    \label{sec:5}
    
    To comprehensively evaluate the performance of VLN-AR, we conduct extensive experiments and ablation studies focusing on four key questions:
    (1) How does VLN-AR compare with state-of-the-art VLN and VLA methods for aerial navigation?
    (2) How well does VLN-AR perform in real-world environments?
    (3) How robust is the proposed VLN-AR across diverse scenarios?
    (4) Are the proposed design components necessary and effective?
    
    To address these questions, Sec.~\ref{sec:benchmark_sim} first presents comprehensive simulation benchmark evaluations against representative VLN and VLA baselines to assess the aerial navigation capability of VLN-AR (Question~(1)). Furthermore, Sec.~\ref{sec:real_world} then conducts extensive real-world experiments to validate the effectiveness and robustness of the proposed framework across diverse environments (Questions~(2) and~(3)). Finally, Sec.~\ref{sec:ablation_analysis} provides detailed ablation studies to quantify the contribution of each key component within VLN-AR (Question~(4)).

    \renewcommand{\arraystretch}{1.4}
    \begin{table}[t]
        \small
        \centering	
        \caption{Hyperparameters for Multi-stage Training.}
        {
            \begin{tabular}{ccc}	
                \hline			
                \textbf{Training Stage}
                & \textbf{Hyperparameter}
                & \textbf{Value}
                \\ \hline	
                \multirow{7}{*}{\makecell{Stage I\\(Grounding-\\Reasoning-\\enhanced \\SFT)}}
                & Batch size & 32 \\ \cline{2-3}
                & Learning rate & $5\times10^{-5}$ \\ \cline{2-3}
                & Epochs & 1 \\ \cline{2-3}
                & Weight decay & 0.01 \\ \cline{2-3}
                & $\lambda_{\text{vqa}}$ & 1.0 \\ \cline{2-3}
                & $\lambda_{\text{bbox}}$ & 1.0 \\ \cline{2-3}
                & $\lambda_{\text{reason}}$ & 0.3 \\  \hline
                
                \multirow{7}{*}{\makecell{Stage II\\(Navigation-\\specific \\SFT)}}
                & Batch size & 16 \\ \cline{2-3}
                & Learning rate & $2\times10^{-5}$ \\ \cline{2-3}
                & Epochs & 2 \\ \cline{2-3}
                & Weight decay & 0.005 \\ \cline{2-3}
                & $\lambda_{\text{pose}}$ & 1.0 \\ \cline{2-3}
                & $\lambda_{\text{finish}}$ & 0.5 \\ \cline{2-3}
                & $\lambda_{\text{replan}}$ & 0.5 \\ \hline
                
                \multirow{5}{*}{\makecell{Stage III\\(Reasoning\\-aware\\Navigation)}}
                & Group size & 16 \\ \cline{2-3}
                & Learning rate & $1\times10^{-5}$ \\ \cline{2-3}
                & Exploration temperature & 0.8 \\ \cline{2-3}
                & Clip ratio & 0.2 \\ \cline{2-3}
                & Reward discount factor & 0.95 \\ \hline
            \end{tabular}
        }
        \label{tab:TrainingHyperparams}	
    \end{table}
    \renewcommand{\arraystretch}{1}

    \setlength{\tabcolsep}{6pt} 
    \renewcommand{\arraystretch}{1.5}
    \begin{table*}[htp]
        \small
    	\centering	
    	\caption{Benchmark comparison of different methods on UAV navigation datasets}
    	{
    		\begin{tabular}{ccccccccccc}	
    			\hline			
                \multirow{2}{*}{\textbf{Types}}
                &\multirow{2}{*}{\textbf{Methods}}
                & \multicolumn{3}{c}{\textbf{AerialVLN Dataset} \cite{liu2023aerialvlnvisionandlanguagenavigationuavs}}
                & \multicolumn{3}{c}{\textbf{OpenFly Dataset} \cite{gao2025openflycomprehensiveplatformaerial}}
                & \multicolumn{3}{c}{\textbf{Our Simulation Dataset}} \\
    
                \cline{3-11}
                &
                & \makecell{\textbf{SR(\%)}$ \uparrow $}
    			& \makecell{\textbf{NE(m)}$ \downarrow $}
                & \makecell{\textbf{TT(s)}$ \downarrow $}
                & \makecell{\textbf{SR}$ \uparrow $}
    			& \makecell{\textbf{NE}$ \downarrow $}
                & \makecell{\textbf{TT}$ \downarrow $}
                & \makecell{\textbf{SR}$ \uparrow $}
    			& \makecell{\textbf{NE}$ \downarrow $}
                & \makecell{\textbf{TT}$ \downarrow $}\\ 
                \hline	

                \multirow{4}{*}{\makecell{\textbf{End-to-end}\\ \textbf{VLN Methods}}}
                & OpenUAV \cite{wang2024realisticuavvisionlanguagenavigation}
    			& 23.02 & 60.11 & 55.76
                & 20.36 & 76.02 & 61.43
                & 16.41 & 14.88 & 25.51\\ \cline{3-11}
                
                & AerialVLN \cite{liu2023aerialvlnvisionandlanguagenavigationuavs}
    			& 33.17 & 59.48 & 46.91
                & 35.73 & 66.44 & 52.26
                & 36.21 & 16.93 & 21.58\\ \cline{3-11}
    
    			& OpenFly \cite{gao2025openflycomprehensiveplatformaerial}
    			& 34.03 & 64.21 & 47.02
                & 30.94 & 77.54 & 48.11
                & 26.73 & 14.14 & 20.02\\ \cline{3-11}
    
    			& UAV-Flow \cite{wang2025uavflowcolosseorealworldbenchmark}
    			& 39.66 & 54.42 & 51.37
                & 38.38 & 72.63 & 52.94
                & 34.18 & 12.77 & 25.11\\ \hline
    
                \multirow{3}{*}{\makecell{\textbf{MLLM-based} \\\textbf{Navigation}\\ \textbf{Agent}}}
                & CityNavAgent \cite{zhang2025citynavagent}
    			& 31.84 & 63.02 & 113.57
                & 21.75 & 74.93 & 128.20
                & 34.96 & 14.60 & 26.58\\ \cline{3-11}
    
                & SPF \cite{hu2025seepointflylearningfree}
    			& 44.16 & 42.81 & 90.63
                & 39.52 & 49.24 & 93.01
                & 42.41 & 13.73 & 21.44\\ \cline{3-11}
                
                & Fly0 \cite{xu2026fly0}
    			& 47.54 & 44.20 & 82.47
                & 49.67 & 43.45 & 78.56
                & 53.16 & 9.41 & 19.47\\ \hline
    
                \multirow{3}{*}{\makecell{\textbf{End-to-end}\\ \textbf{VLA Methods}}}
                & OpenVLA \cite{kim2024openvlaopensourcevisionlanguageactionmodel}
    			& 29.76 & 55.90 & 56.02
                & 24.19 & 70.48 & 62.19
                & 25.88 & 20.72 & 27.64\\  \cline{3-11}
    
                & $\pi_{0}$ \cite{black2024pi0visionlanguageactionflowmodel}
    			& 37.95 & 47.26 & 54.41
                & 41.63 & 52.04 & 51.77
                & 33.82 & 13.41 & 25.02\\  \cline{3-11}
    
                & GR00T N1.5 \cite{nvidia2025groot}
    			& 39.12 & 41.73 & 51.55
                & 35.41 & 42.28 & 49.33
                & 40.18 & 10.92 & 23.35\\ \hline
    
                \multirow{3}{*}{\makecell{\textbf{Zero-shot}\\ \textbf{MLLM-based}\\ \textbf{Navigation}}}
                & Qwen3-VL-4B \cite{bai2025qwen3}
    			& 15.34 & 73.25 & 72.51
                & 14.37 & 82.31 & 81.34
                & 29.24 & 17.72 & 32.31\\  \cline{3-11}
    
                & ChatGPT 5.2 \cite{openai2025chatgpt}
    			& 21.12 & 62.47 & 61.05
                & 19.13 & 68.38 & 67.66
                & 36.32 & 14.24 & 25.04\\  \cline{3-11}
  
                & Gemini 3 Pro \cite{google2025gemini}
    			& 31.61 & 55.46 & 57.98
                & 37.42 & 50.37 & 52.51
                & 42.88 & 13.72 & 24.64\\  \hline
    
                \multirow{2}{*}{\makecell{\textbf{Hierarchical VLN}}} 
                & \add{VLA-AN\cite{wu2025vla}}
    			& \add{67.37} & \add{31.24} & \add{50.95}
                & \add{64.67} & \add{33.09} & \add{47.18}
                & \add{87.42} & \add{7.56} & \add{17.73} \\ \cline{3-11}
                & \textbf{Ours}
    			& \textbf{78.15} & \textbf{22.12} & \textbf{41.51}
                & \textbf{76.88} & \textbf{23.22} & \textbf{40.07}
                & \textbf{92.66} & \textbf{5.75} & \textbf{16.18} 
                \\ \hline
    		\end{tabular}
    	}
    	\label{tab:benchmark}	
    \end{table*}
    \renewcommand{\arraystretch}{1}
        
    \begin{figure}[t]
      \centering
      \includegraphics[width=0.9\linewidth]{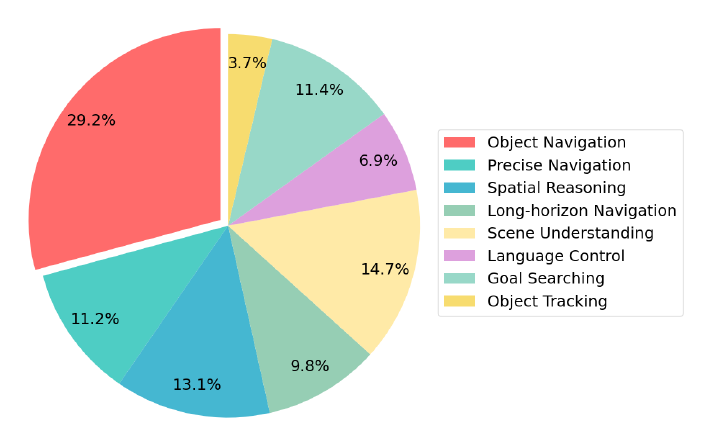}
      \caption{Task distribution of test data.}
      \label{fig:test_data}
      \vspace{-0.3cm}
    \end{figure}

    \subsection{Experimental Setup}
    \label{sec:setup}
    This study systematically evaluates the proposed method in both simulated and real-world environments. The simulation experiments are first conducted using open UAV navigation datasets, including AerialVLN \cite{liu2023aerialvlnvisionandlanguagenavigationuavs} and OpenFly \cite{gao2025openflycomprehensiveplatformaerial}, which are built upon the Unity Engine and AirSim platforms. Importantly, these datasets are not used during our training stage. These datasets primarily target \underline{\textbf{outdoor}} UAV navigation tasks and contain different urban and suburban scene maps, along with more than 14K manually annotated natural-language navigation instructions.
    
    To validate aerial navigation in more complex indoor scenarios, we also construct a 3DGS-based \underline{\textbf{indoor}} simulation dataset based on the Unity Engine, covering a variety of environments such as office scenes, family settings, shopping malls, and cinemas. This simulation dataset contains approximately 10K navigation samples, which are unseen during the training stage. Based on these scenes, we design several representative navigation task categories, including object navigation (e.g., “Head to the white circular column from the corridor”), precise navigation (e.g., “Go to the right of the chair on the left”), spatial reasoning (e.g., “Find the third tree from near to far”), language control (e.g., “move forward 10 meters”), long-horizon navigation, which refers to instructions containing three or more sequential sub-tasks (e.g., “Head to the black chair, then the front circular pillar, and finally land on the carpet”), scene understanding (e.g., “Navigate to the device capable of regulating the temperature.”), goal searching (e.g., “Search for the blue bag in a cluttered environment.”), and object tracking (e.g., “Continuously track the person wearing jeans”), etc. 
    The proportions of different task types in the evaluation set are also reported in Fig.~\ref{fig:test_data}. \add{We will also make these test sets open-source as soon as possible to serve as a future VLN baseline for the robotics community.}
    
    We adopt standard evaluation metrics widely used in UAV VLN research to quantitatively evaluate the performance of different approaches. The evaluation metrics include:
    
    \begin{itemize}
        \item \textbf{Success Rate (SR):} the percentage of trials in which the UAV finishes the task without collision, within a distance threshold to the goal (5 m for outdoor AerialVLN and OpenFly datasets and 1 m for our indoor simulation dataset). \addagain{For long-horizon tasks involving multiple sequential subtasks, an episode is considered successful only if all subtasks are completed successfully.}
        \item \textbf{Navigation Error (NE):} the average Euclidean distance between the final UAV position and the target location (in meters). \addagain{For long-horizon tasks, NE is computed for each subtask and accumulated across subtasks.}
        \item \textbf{Trajectory Time (TT):} the average time required to successfully complete a navigation task (in seconds), which reflects both navigation efficiency and path rationality.
    \end{itemize}
    
    As described in Sec.~\ref{sec:Nav_model}, existing UAV navigation foundation models encompass multiple paradigms, and we select the following four categories for further comparison:

    \begin{itemize}
        \item \textbf{End-to-end VLN Methods:} The first category includes UAV VLN methods such as OpenFly \cite{gao2025openflycomprehensiveplatformaerial}, AerialVLN \cite{liu2023aerialvlnvisionandlanguagenavigationuavs}, OpenUAV \cite{wang2024realisticuavvisionlanguagenavigation}, and UAV-Flow \cite{wang2025uavflowcolosseorealworldbenchmark}, which typically adopt end-to-end policies that directly map visual inputs to flight commands.
        \item \textbf{MLLM-based Navigation Agent:} The second category consists of zero-shot navigation agents, such as CityNavAgent \cite{zhang2025citynavagent} and SPF \cite{hu2025seepointflylearningfree}, which rely on pretrained foundation models for semantic grounding and reasoning, and then convert detected targets into goal waypoints via rule-based modules.
        \item \textbf{End-to-end VLA Methods:} The third category includes general VLA models, such as OpenVLA \cite{kim2024openvlaopensourcevisionlanguageactionmodel}, $\pi_{0}$ \cite{black2024pi0visionlanguageactionflowmodel}, and GR00T N1.5 \cite{nvidia2025groot}. Although not designed specifically for navigation, these models exhibit strong application potential and cross-domain transferability. \add{For OpenVLA and $\pi_0$, we mainly follow the training setup and model versions of UAV-FLOW~\cite{wang2025uavflowcolosseorealworldbenchmark}. We fine-tune GR00T N1.5 on the same data using a largely consistent configuration. And each model takes image observations and a navigation prompt as input and outputs a flight trajectory.} 
        \item \textbf{Zero-shot MLLM-based Navigation:} The fourth category directly leverages pretrained multimodal foundation models \cite{bai2025qwen3, openai2025chatgpt,google2025gemini} for zero-shot UAV navigation. Although not designed for navigation, we evaluate their zero-shot performance to probe the capability limits of current MLLMs in embodied aerial navigation. \add{For the zero-shot MLLM baselines, we keep all model parameters frozen and evaluate them through API calls.}
    \end{itemize}
    
    Training parameters of VLN-AR are provided in Tab.~\ref{tab:TrainingHyperparams}.

    \begin{figure*}
      \centering
      \includegraphics[width=0.82\linewidth]{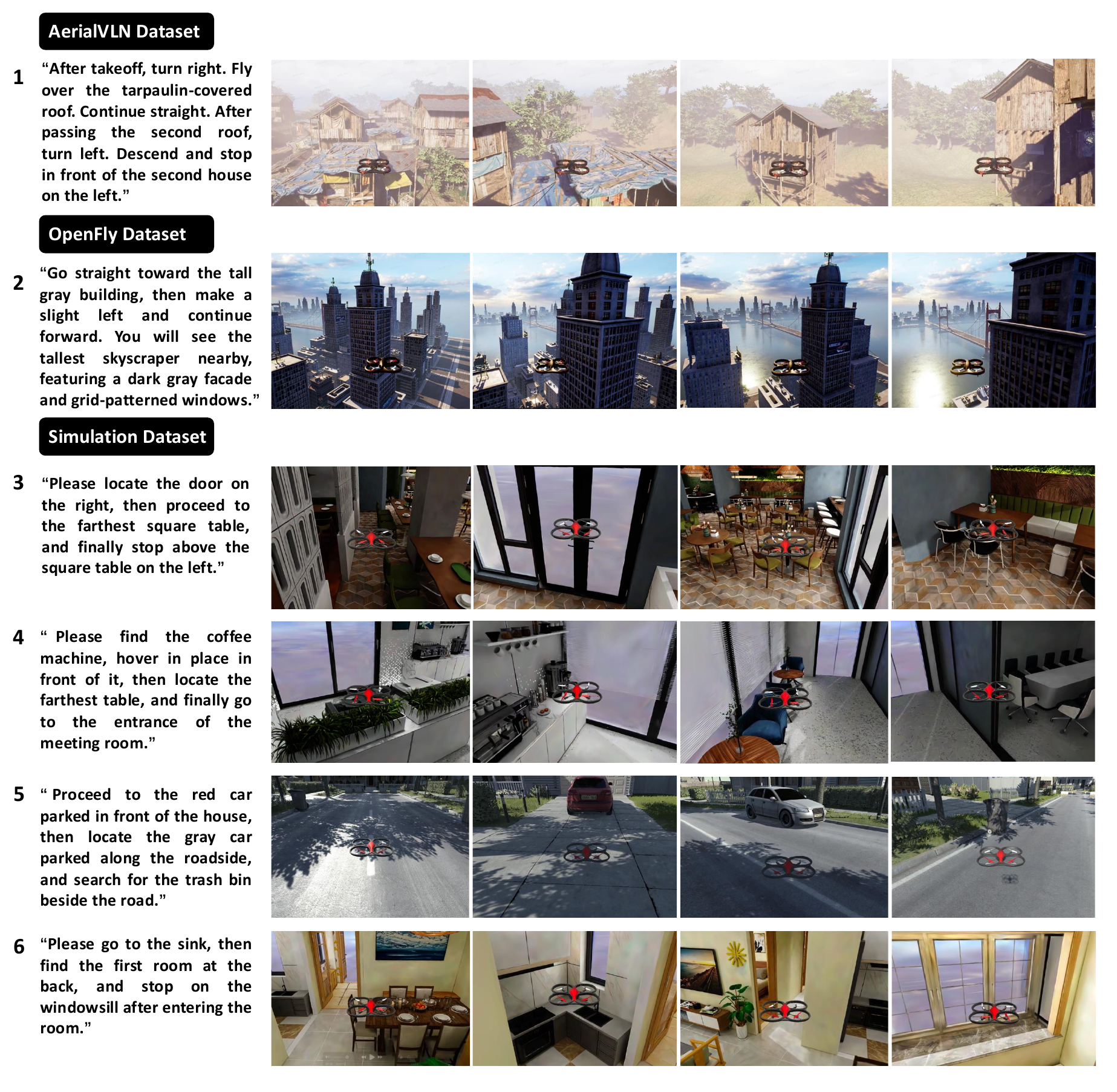}
      \caption{Simulation tests of VLN-AR across different navigation datasets using both visual and language inputs.}
      \label{fig:simulation}
      \vspace{-0.2cm}
    \end{figure*}

    \subsection{Benchmark Results and Analysis}
    \label{sec:benchmark_sim}
    We conduct comparative experiments across the above four representative categories of approaches. More typical simulation results are shown in Fig.~\ref{fig:simulation}. The proposed method achieves the highest SR across all different datasets, while also demonstrating superior performance in NE and TT, as shown in Tab.~\ref{tab:benchmark}. This improvement primarily stems from the proposed hierarchical framework, which avoids blindly adopting a fully end-to-end paradigm under limited training data. By jointly optimizing multi-stage SFT and RFT training, the model acquires stronger capabilities across diverse navigation tasks. In particular, the introduction of the agent module further enables task decomposition and memory maintenance, substantially improving the success rate of long-horizon navigation tasks.
    Furthermore, we observe that AerialVLN and OpenFly datasets often contain ambiguous instructions corresponding to multiple plausible targets within a scene. For example, “navigate to the white building near the river” may match several objects, while only one is annotated as correct. As a result, most methods achieve lower overall success rates on these datasets than on our simulation dataset. 
    
    \subsubsection{Comparison with End-to-end VLN Methods} 
    UAV VLN methods typically use black-box neural policies to map visual observations to flight commands. On OpenFly, they achieve only 20\%–40\% SR, while ours reaches 76.88\%.
    This gap is particularly evident under complex semantic instructions. For example, the instruction “proceed to the large beige and brown building with windows” requires both object recognition and fine-grained language grounding aligned with visual scenes. However, existing methods, lacking explicit semantic and spatial modeling, often fail to achieve reliable cross-modal alignment, limiting their planning and generalization abilities.
    The problem is further exacerbated when training data is much smaller than the pretraining scale of general large foundation models, leading to degraded transfer performance.    
    In contrast, our method consistently outperforms prior approaches across datasets. The results demonstrate that explicit 3D geometric pose representations and enhanced language reasoning effectively address the limitations of current UAV VLN methods in spatial modeling and semantic grounding, improving overall navigation performance and generalization, as shown in Fig.~\ref{fig:simulation}.
    
    \subsubsection{Comparison with MLLM-based Navigation Agent} 
    These MLLM-based navigation agents are typically zero-shot methods that rely on pretrained foundation models for step-by-step reasoning and decision-making. They usually first predict 2D image-space waypoints and then lift them to 3D using depth maps. This strategy fails in free-space or non-physical target cases, such as “pass through the circular frame” or “move around the tree from the right side,” where depth cannot provide reliable guidance of the goal position. As a result, NE increases significantly and performance degrades.
    For example, CityNavAgent, SPF, and Fly0 achieve only 31.84\%, 44.16\%, and 47.54\% SR on AerialVLN, with further drops on our dataset containing more free-space targets. In contrast, our method can directly predict 3D waypoints and yaws, avoiding absolute reliance on depth maps, enabling stable execution and better orientation.
    Moreover, these methods lack motion planning that accounts for UAV dynamic constraints, resulting in unstable navigation behavior, lower SR, and longer TT. In contrast, our action expert explicitly models UAV dynamics and collision constraints, and incorporates trajectory time optimization, thereby significantly reducing collision risk and improving navigation efficiency.

    \subsubsection{Comparison with End-to-end VLA Methods} 
    The proposed method consistently outperforms end-to-end VLA approaches across all datasets. On AerialVLN, it achieves 78.15\% SR, compared to 29.76\%, 37.95\%, and 39.12\% for OpenVLA, $\pi_{0}$, and GR00T N1.5, respectively, while also reducing NE to 22.12 m. Similar gaps are observed on other datasets.
    This advantage arises from limitations of end-to-end VLA methods, which directly predict low-level controls and must jointly learn language understanding, perception, and control. Under limited data and large model sizes, this leads to poor generalization and unstable trajectories, especially in complex scenarios.
    In contrast, our method predicts 3D pose commands, simplifying learning and decoupling high-level reasoning from motion execution. This results in more stable trajectories, improved accuracy, and better handling of complex instructions and dynamic environments.

    \subsubsection{Comparison with Zero-shot MLLM-based Navigation} 
    For a relatively fair comparison, we adopt a consistent workflow: the visual observations and language instruction are provided as inputs, and the models are required to predict the target 3D poses \add{and yaw angles in the UAV body frame}, which are \add{parsed using the same format and} passed to \add{the same} action expert for trajectory generation. 
    As shown in Tab.~\ref{tab:benchmark}, they demonstrate moderate performance across datasets but remain significantly inferior to our method. For instance, on AerialVLN, their SR values are 15.34\%, 21.12\%, and 31.61\%, respectively, far below our 78.15\%; similar trends are observed on OpenFly. Meanwhile, their NE is consistently higher, indicating lower navigation accuracy.
    This gap primarily arises from their reliance on general-purpose multimodal reasoning, without task-specific spatial modeling or explicit constraints, making precise navigation decisions difficult. In complex or long-horizon tasks, this often results in unstable trajectories and inefficient paths.
    These results demonstrate that existing MLLMs, benefiting from large-scale pretraining, have already acquired preliminary navigation capabilities. However, there remains a significant gap in general-purpose navigation performance, with substantial room for improvement in success rate, localization accuracy, and robustness.
    
    \add{Furthermore, the previous work VLA-AN~\cite{wu2025vla} shows a success rate more than 10 percentage points lower on the AerialVLN and OpenFly datasets, possibly because these benchmarks contain more long-horizon tasks, for which our Agent mechanism, finer-grained design, and more diverse training data provide greater benefits. On our self-built dataset, where long-horizon tasks account for a smaller proportion, the performance gap is correspondingly reduced.}

    \subsection{Real-World Results}
    \label{sec:real_world}
    \begin{figure}[t]
      \centering
      \includegraphics[width=0.88\linewidth]{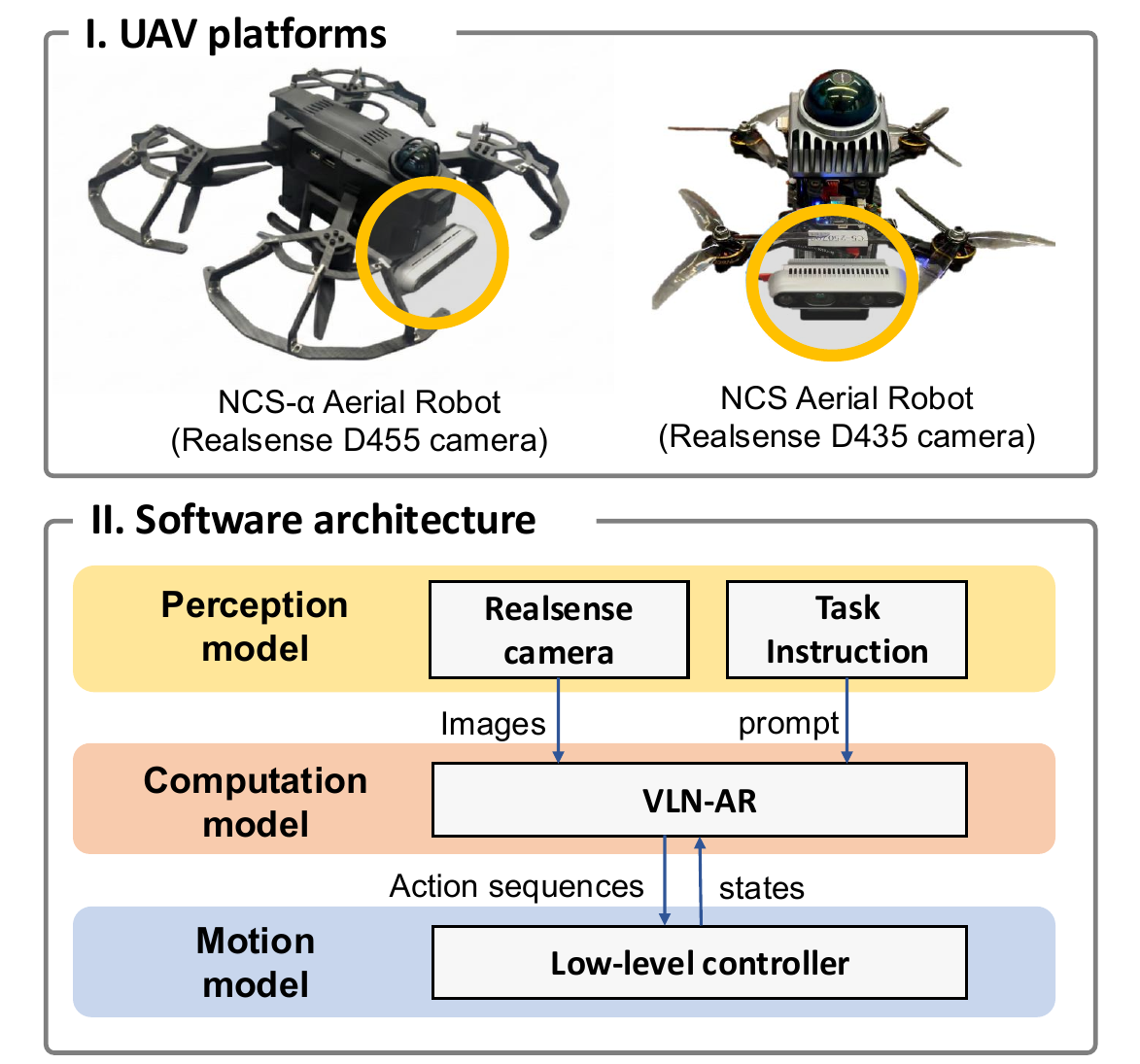}
      \caption{Hardware and software architectures of two different experimental drones.}
      \label{fig:platform}
    \end{figure}
    
    \begin{figure*}[htp]
      \centering
      \includegraphics[width=0.87\linewidth]{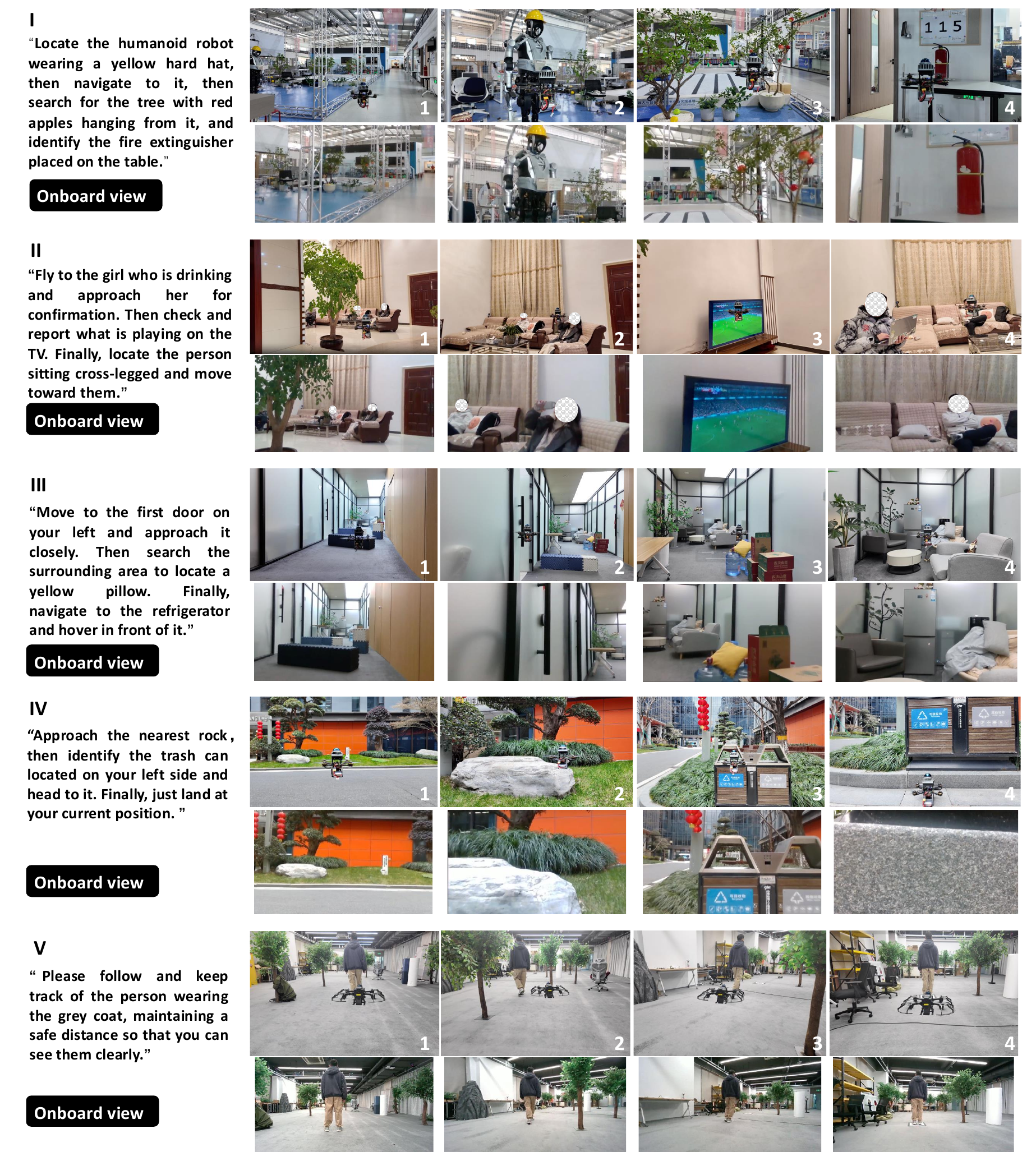}
      \caption{Real-world experiments for navigation scenarios utilizing both visual and language inputs.}
      \label{fig:multi_scene}
    \end{figure*}

    Beyond simulation, we conduct extensive real-world experiments to validate the proposed VLN-AR method. We employ two UAV platforms—a compact Differential Robotics NCS aerial robot and a larger NCS-$\alpha$ aerial robot—both equipped with an Intel RealSense camera for visual perception. Each platform is also equipped with an NVIDIA Jetson Orin NX platform for fully onboard model inference and reasoning. This lightweight module (about 80 g) provides approximately 100 TOPS of compute, making it suitable for small-payload aerial systems. The overall system enables efficient onboard perception and control, while achieving strong generalization across diverse real-world scenarios, as detailed below. The hardware and software architectures are illustrated in Fig.~\ref{fig:platform}. In real-world evaluation, \add{the hardware-task prompts also follow the same task definitions and success criteria as those used in simulation.} 
    
    \subsubsection{Real-world Performance}
    We evaluate the proposed method on over 100 diverse real-world test cases to assess its practical effectiveness across a wide range of capabilities, including the aforementioned \add{(30\%)} object navigation, \add{(20\%)} spatial grounding, \add{(20\%)} scene reasoning, \add{(10\%)} long-horizon planning, \add{(10\%)} target tracking, and \add{(10\%)} more.
    \add{Our task scenarios cover logistics facilities, home environments, office buildings, outdoor parks, testing grounds, shopping districts, and others, as illustrated in Fig.~10.}
    The system is deployed on both UAV platforms, where observations are asynchronously transmitted while previously generated actions are continuously executed, enabling smooth closed-loop control. The overall success rate reaches 90.1\% (91/101), which is comparable to the simulation results, demonstrating strong real-world generalization capability. Furthermore, for long-horizon instructions such as \textit{“Move to the first door on your left and approach it closely. Then search the surrounding area to locate a yellow pillow. Finally, navigate to the refrigerator and hover in front of it,”} our method successfully decomposes sequential sub-tasks and executes them reliably.
    \add{As mentioned above, the proposed action expert provides a safety fallback for collision avoidance, and no experimental failures are attributed to collisions. Most failure cases arise from an inability to accurately reach the target poses. For example, for the instruction “navigate to the fifth tree from near to far among multiple trees,” the model cannot precisely interpret the attribute “the fifth from near to far.” Targets in instructions such as “navigate to the spoon in the kitchen” are too small to be reliably recognized and understood from visual observations. In addition, image blur introduced by the RealSense D435 camera during UAV motion can also lead to task failures.}
    
    \subsubsection{Onboard Inference Performance}
    As described in Sec.~\ref{sec:onboard}, we introduce a series of systematic optimizations to improve onboard inference efficiency. We further validate the proposed approach through real-world experiments.
    As shown in Tab.~\ref{tab:onboard_interference}, the optimized decoding pipeline consistently achieves substantial latency reductions across all model scales compared to the unoptimized baseline. In particular, the 4B-AWQ model requires only $\sim$946.7 GFLOPs while achieving a decoding latency of 0.062 s/token, representing a significant improvement over its baseline counterpart (0.508 s/token).
    We further evaluate models of different scales under the 30 W performance mode. The 2B model achieves the lowest latency of 0.032 s/token (vs. 0.227 s/token baseline), while the 7B model, despite its higher computational cost ($\sim$1815.8 GFLOPs), maintains a practical latency of 0.110 s/token (vs. 1.034 s/token baseline). Overall, our method yields a consistent 7.1$\times$–9.4$\times$ speedup across all configurations.
    These gains stem from the proposed system-level optimizations (Sec.~\ref{sec:onboard}), which effectively reduce decoding overhead and improve hardware utilization. During closed-loop navigation, the integration of asynchronous perception and a continuous action expert further enables high-frequency, real-time command generation, which is crucial for agile flight in complex and dynamic environments.

    \setlength{\tabcolsep}{5pt} 
    \renewcommand{\arraystretch}{1.5}
    \begin{table}[tp]
    	\centering	
    	\caption{Onboard inference performance on different models}
    	{
    		\begin{tabular}{cccc}	
    			\hline			
                \multirow{2}{*}{\textbf{Model}}
                & \multirow{2}{*}{\textbf{\makecell{Floating Point \\ Operations}}} 
                & \multirow{2}{*}{\textbf{\makecell{Baseline Decode\\ Speed (s/token)}}} 
                & \multirow{2}{*}{\textbf{\makecell{Our Decode\\ Speed  (s/token)}}} \\
                \\
                \hline	
    			7B-AWQ & $\sim$$~1815.8$ GFLOPs & 1.034 & \textbf{0.110} \\  \hline	
                4B-AWQ & $\sim$$~946.7$ GFLOPs  & 0.508 & \textbf{0.062} \\ \hline		
    			2B-AWQ & $\sim$$~563.3$ GFLOPs  & 0.227 & \textbf{0.032} \\ \hline
    		\end{tabular}
    	}
    	\label{tab:onboard_interference}
    \end{table}
    \renewcommand{\arraystretch}{1}  

    \begin{figure*}[ht]
      \centering
      \includegraphics[width=0.82\linewidth]{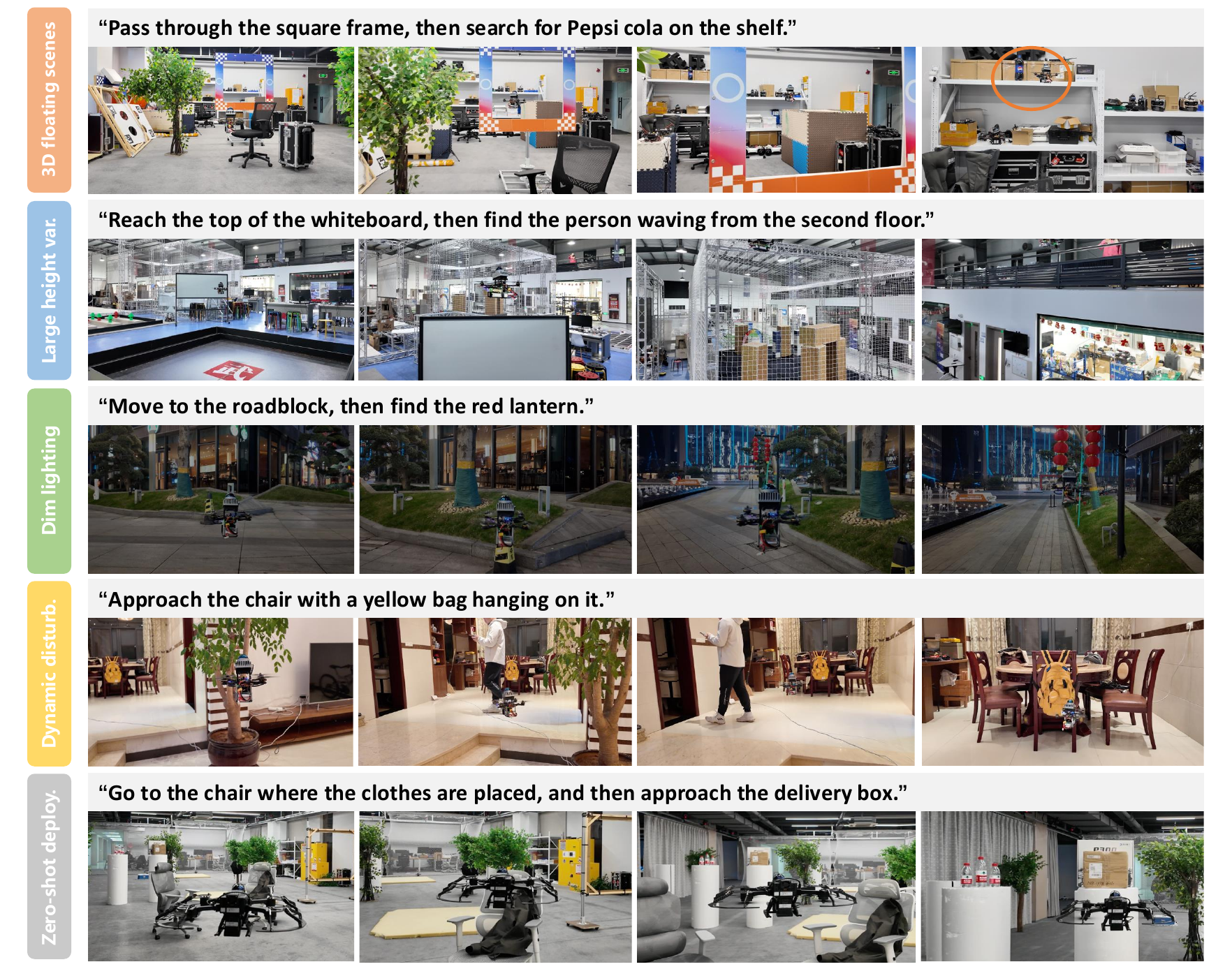}
      \caption{Robustness tests on 3D floating scenes, large height variations, dim lighting, dynamic disturbance, and zero-shot deployment.}
      \label{fig:extreme_test}
      \vspace{-0.1cm}
    \end{figure*}
    
    \subsubsection{Robustness Test}
    To validate the environmental adaptability and anti-interference capability of VLN-AR in complex real-world scenarios, we design a series of challenging robustness test tasks in 3D aerial navigation, including 3D floating scenes, large height variations, dim lighting environments, dynamic disturbance, and zero-shot deployment in unseen scenes. All test tasks adopt open-vocabulary natural language instructions without additional fine-tuning or parameter adjustment for the model, directly verifying the generalization and robustness of the VLN-AR framework. The results from real-world experiments are summarized in Fig.~\ref{fig:extreme_test} and Tab.~\ref{tab:Robustness}, as detailed below:

    \begin{itemize}

        \item \textbf{3D floating scene test.} We construct a navigation environment with suspended objects and discontinuous spatial structures, \add{using} instructions such as ``Pass through the square frame, then search for Pepsi cola on the shelf''. The results show that VLN-AR can accurately perceive the 3D spatial layout through multi-frame visual fusion and dynamic memory buffering, and generate safe 3D poses for obstacle avoidance and target search, \add{despite} discontinuous scene geometry.

        \item \textbf{Large height variation test.} The navigation tasks involve vertical \add{transitions exceeding} 5 meters, such as ``Reach the top of the whiteboard, then find the person waving from the second floor''. VLN-AR decouples horizontal and vertical motion planning through its structured 3D action representation, while the geometric safety correction module dynamically adjusts the vertical trajectory \add{using depth information to avoid collisions and complete height-crossing navigation safely}.
        
        \item \textbf{Dim lighting environment test.} As shown in Fig.~\ref{fig:extreme_test}, we conduct experiments in outdoor night scenes with only ambient light. The vision encoder of VLN-AR combines dynamic token selection to extract valid visual features from low-quality images, while the LLM reasoning module supplements environmental context through historical memory, \add{thereby maintaining reliable spatial grounding, navigation decisions, and} stable task execution under insufficient lighting.

    \setlength{\tabcolsep}{10pt} 
    \renewcommand{\arraystretch}{1.5}
    \begin{table}[t]
        \small
    	\centering	
    	\caption{Robustness test of different navigation scenes}
    	{
    		\begin{tabular}{cc}	
    			\hline			
    			\textbf{Test scenes}
                & \makecell{\textbf{Results}}
                \\ \hline	

                3D floating scene
                & 18 / 20 \\   \hline	
                
                Large height variation
    			& 19 / 20 \\  \hline	

                Dim lighting environment
    			& 17 / 20 \\ \hline
                
    			Dynamic obstacle interference
    			& 17 / 20 \\ \hline
    
                Zero-shot deployment
    			& 18 / 20
                \\ \hline
    		\end{tabular}
    	}
    	\label{tab:Robustness}	
    \end{table}
    \renewcommand{\arraystretch}{1}

    

        \item \textbf{Dynamic obstacle interference test.} We introduce unexpected human \add{disturbances} during navigation, where a person suddenly enters the UAV’s flight path from the side. As shown in Fig.~\ref{fig:extreme_test}, the multi-frame temporal fusion mechanism quickly detects the moving obstacle, and the geometric safety correction module adjusts intermediate 3D waypoints. Meanwhile, the structured waypoint abstraction keeps the high-level language-conditioned plan stable, allowing the low-level controller to \add{locally deform the trajectory} without interrupting the task objective. As a result, VLN-AR smoothly detours around the moving person and continues the original instruction, demonstrating strong robustness and safety under dynamic disturbances.

        \item \textbf{Zero-shot deployment test.} We select completely unseen real-world scenes \add{with no scene elements appearing in the training set} and use navigation instructions such as ``Move to the roadblock, then find the red lantern'' and ``Go to the chair where the clothes are placed, and then find the delivery box''. Without any fine-tuning, VLN-AR relies on its \add{three-stage training-induced} general spatial reasoning capability and cognitive-driven task decomposition mechanism to autonomously complete \add{these tasks with a 90\% success rate}, as shown in Tab.~\ref{tab:Robustness}.
    \end{itemize}

    \subsubsection{Limitations in Real-world Experiments}
    Despite the overall robustness of the system, the real-world experiments reveal several limitations. First, in scenes containing many visually similar objects, ambiguous language instructions may cause the semantic model to localize an object that resembles the target but is not the actual goal, resulting in navigation deviations. Second, most failure cases stem from inaccurate spatial estimation, which may occur at the interface between semantic understanding and 3D geometric mapping. For distant structures or small targets, the predicted goal point may suffer from the 2D-to-3D projection process, which can lead to coordinate deviations and cause the UAV to either undershoot or overshoot the target location, thereby reducing the success rate.

    \setlength{\tabcolsep}{12pt} 
    \renewcommand{\arraystretch}{1.5}
    \begin{table*}[h]
        \centering
        \small
        \setlength{\tabcolsep}{3pt}
        \caption{Ablation study of the multi-stage training strategy (Success rates \%)}
        {
            \begin{tabular}{cccccccc}
                \hline
                \multirow{2}{*}{\textbf{Methods}}
                 & \multirow{2}{*}{\makecell{\textbf{Grounding-reasoning} \\ \textbf{-enhanced SFT}}} & \multirow{2}{*}{\makecell{\textbf{Navigation} \\ \textbf{-specific SFT}}} & 
                \multirow{2}{*}{\makecell{\textbf{RFT for Reasoning} \\ \textbf{-aware Navigation}}}
                & \multirow{2}{*}{\makecell{\textbf{Scene}\\\textbf{Understanding}}}
                & \multirow{2}{*}{\makecell{\textbf{Spatial}\\\textbf{Reasoning}}}
                & \multirow{2}{*}{\makecell{\textbf{Precise}\\\textbf{Navigation}}}
                & \multirow{2}{*}{\makecell{\textbf{Long-horizon}\\\textbf{Navigation}}} \\ 
                & & & & & & \\ \hline
                Baseline & $\times$ & $\times$ & $\times$ & 77.4 & 68.2 & 21.3 & 17.8 \\ \hline
                Stage I & $\checkmark$ & $\times$ & $\times$ & \textbf{87.2} & 94.5 & 28.1 & 22.5 \\ \hline
                Stage II & $\times$ & $\checkmark$ & $\times$ & 75.6 & 64.3 & 68.5 & 64.4  \\ \hline
                Stage I + Stage II & $\checkmark$ & $\checkmark$ & $\times$  & 80.2 & 88.3 & 60.6 & 61.5 \\ \hline
                Stage III & $\times$ & $\times$ & $\checkmark$ & 82.6 & 76.1 & 45.4 & 36.3 \\ \hline
                \textbf{Ours} & $\checkmark$ & $\checkmark$ & $\checkmark$ & 85.5 & \textbf{94.6} & \textbf{85.7} & \textbf{88.0} \\ \hline
            \end{tabular}
        }
        \label{tab:training}
    \end{table*}
    \renewcommand{\arraystretch}{1}

    \subsection{Ablation Studies}
    \label{sec:ablation_analysis}    

    \begin{figure}[t]
      \centering
      \includegraphics[width=0.88\linewidth]{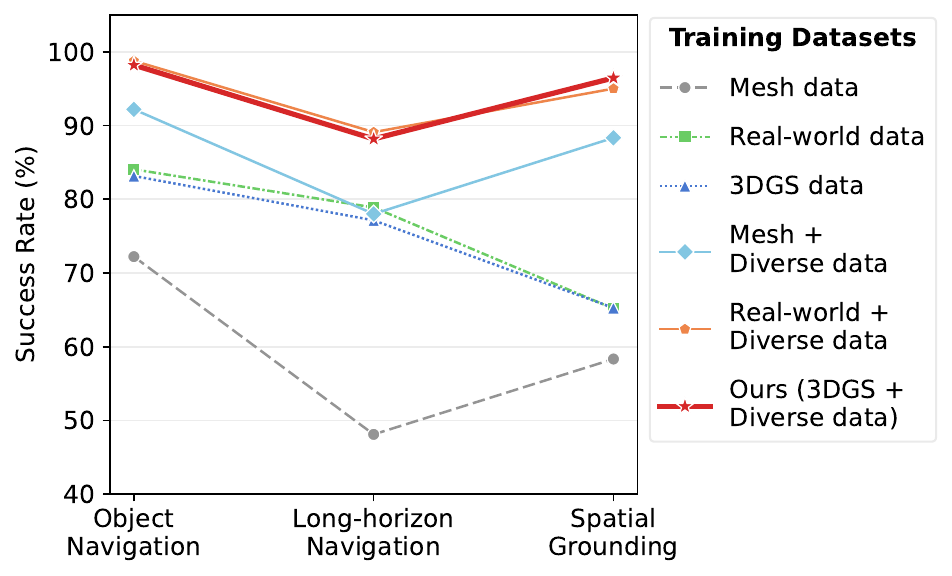}
      \caption{Ablation study of different training navigation datasets.}
      \label{fig:data_comparasion}
      \vspace{-0.2cm}
    \end{figure}

    \subsubsection{Different Navigation Datasets}
    We systematically evaluated the impact of different training datasets on the performance of VLN-AR, including mesh data, real-world scene data, 3DGS data, as well as their combinations with diverse data (grounding, reasoning, spatial VQA, etc.), as shown in Fig.~\ref{fig:data_comparasion}.
    The model trained solely on mesh data exhibited the lowest overall performance, achieving a target navigation success rate of 72.2\%, long-horizon navigation of 48.1\%, and spatial grounding of 58.3\%. This indicates that, although mesh data provides extensive viewpoint coverage, its lack of realism and semantic diversity limits the model’s generalization ability in complex navigation scenarios.    
    In contrast, incorporating real-world scene data leads to significant improvements across all tasks. However, real-world data collection is highly labor-intensive, costly, and difficult to scale, which fundamentally limits its applicability for large-scale training. Despite this constraint, real images enhance the model’s perception of realism and semantic understanding, although multi-step planning remains insufficient for long-horizon navigation. Impressively, as shown in Fig.~\ref{fig:data_comparasion}, 3DGS data achieves similar performance relative to real data, while being significantly easier to collect and fully scalable through automated reconstruction pipelines. \addagain{These results suggest that, under the evaluated experimental settings, high-quality 3DGS-rendered images can achieve performance comparable to real-world data, demonstrating their potential as an alternative training data source. However, this finding may depend on the scene distributions represented by the respective datasets.}
    
    Further combining mesh, real-world, or 3DGS data with diverse data substantially improves performance across all tasks. \add{Each dataset contains approximately 1M samples, while all other training settings, including model initialization, optimization steps, and evaluation protocols, are kept identical. The task distribution comprises navigation data (30\%), general VQA (15\%), spatial VQA (20\%), grounding (15\%), reasoning (10\%), and more (10\%).} \add{T}he long-horizon navigation success rate increases to 78.0\% for mesh + diverse data, reaches 88.7\% for real-world + diverse data, and enhances 88.0\% for 3DGS + diverse data (Ours). These results indicate that diverse data can effectively enhance the model’s spatial reasoning, task generalization, and overall understanding, significantly improving general navigation and spatial grounding.

    \subsubsection{Multi-stage Training Strategy}
    To validate the effectiveness of the proposed progressive three-stage training framework for multi-task aerial navigation, we conduct systematic ablation studies. The results indicate that different training stages contribute distinctly to various task performances (Tab.~\ref{tab:training}). \add{It is worth noting that these results also include the cognitive-driven agent to ensure fairness.} Stage I significantly enhances scene understanding and spatial reasoning, achieving 87.2\% and 94.5\% SR, respectively. These results demonstrate that the grounding-reasoning-enhanced SFT effectively strengthens the model’s visual grounding and high-level reasoning abilities.
    In contrast, Stage II achieves substantially better performance on precise and long-horizon navigation tasks, reaching 68.5\% and 64.4\% SR, respectively. This suggests that navigation-specific training effectively improves trajectory planning and low-level action execution.
    
    Interestingly, jointly applying Stage I and Stage II improves all task categories relative to the baseline, demonstrating complementary benefits between reasoning and navigation learning. However, the joint model does not outperform the corresponding single-stage variants on their specialized tasks, which we attribute to Task Interference~\cite{yu2020gradient} during multi-task optimization. Specifically, optimization competition and representational sharing across heterogeneous objectives reduce task-specific specialization, despite improving the overall generalization capability of the model.
    Finally, incorporating Stage III RFT further enhances the coordination between high-level reasoning and low-level navigation behaviors. \add{During RFT, the policy is equipped with agent mechanisms such as task decomposition, memory fusion, and CoT update, according to the actual training configuration, while all other components are kept fixed.} Combined with the preceding SFT stages, the full model (Ours) achieves the best overall performance across nearly all evaluation metrics, particularly on long-horizon navigation tasks, demonstrating the effectiveness of the proposed progressive training paradigm.

    \begin{figure}[t]
      \centering
      \includegraphics[width=0.82\linewidth]{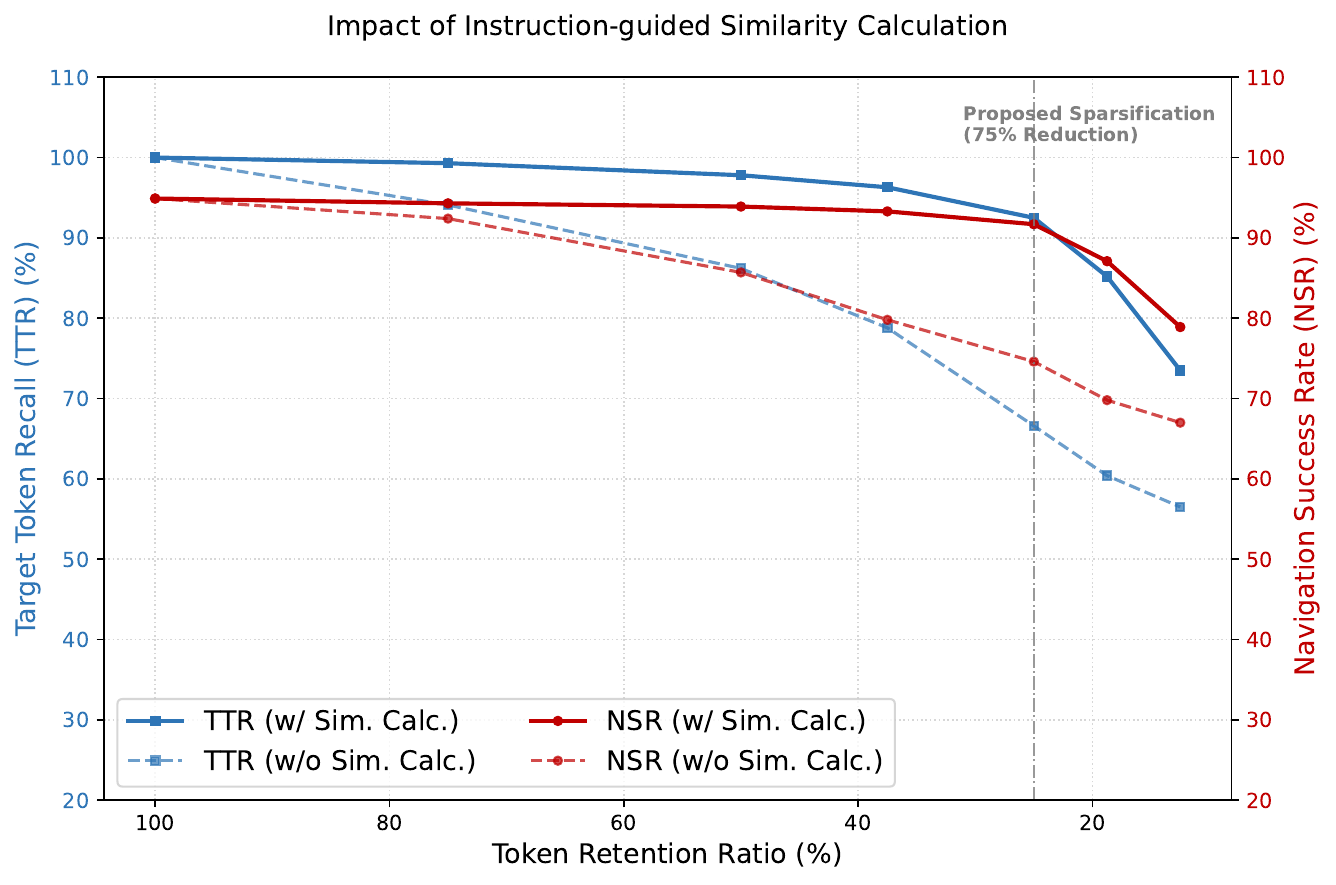}
      \caption{Impact of different token retention ratios for ViT.}
      \label{fig:vit_analysis}
      \vspace{-0.3cm}
    \end{figure}

    \subsubsection{Analysis of ViT Token Sparsification}
    \label{sec:vit_analysis}
    To investigate the impact of the similarity-based filtering mechanism, we compare our approach against a baseline that performs token reduction without our similarity calculation module. As illustrated in Fig.~\ref{fig:vit_analysis}, the experimental results demonstrate that without instruction guidance, both the Target Token Recall (TTR) and Navigation Success Rate (NSR) exhibit distinct degradation as the token retention ratio decreases. This suggests that task-critical spatial semantics—essential for identifying navigation goals—are inadvertently discarded when the model lacks cross-modal task awareness. In contrast, our instruction-guided mechanism maintains a high-fidelity performance plateau ($TTR \approx 92.5\%$, $NSR \approx 91.7\%$) even at an aggressive sparsification level of approximately $25\%$ token retention.     
    Furthermore, the curves reveal a clear theoretical inflection point near a $25\%$ retention ratio. Above this boundary, the cross-modal similarity effectively anchors navigation-critical entropy by filtering task-irrelevant regions (e.g., sky and repetitive textures as seen in outdoor navigation tasks). Below this threshold, the information bottleneck constraints lead to irreversible semantic loss. These findings confirm that our strategy acts as a semantic denoiser rather than a destructive compressor, achieving a superior trade-off between computational efficiency on the constrained computation vehicle and aerial navigation accuracy. Specifically, in extreme scenarios where targets are visually subtle, our method ensures robust semantic preservation where unguided pruning fails catastrophically.

    \subsubsection{Aerial Navigation Agent for VLN-AR}
    To quantitatively assess the contribution of the aerial navigation agent in VLN-AR, we conduct ablation experiments comparing the full model with a variant without the agent. As summarized in Tab.~\ref{tab:agent_ablation}, the inclusion of the agent consistently improves performance across all evaluated tasks. Specifically, object navigation success increases from 95.7\% to 98.1\%, spatial reasoning improves from 92.4\% to 94.6\%, and long-horizon navigation shows the most pronounced gain, rising from 57.4\% to 88.0\%. 
    These results indicate that the agent plays a critical role in enabling robust and precise task planning and decision making, particularly in tasks requiring multi-step reasoning and extended trajectory planning. 
    \addagain{Notably, long-horizon task success also depends on the task-completion indicator $y_t^{\text{finish}}$. We manually inspect 976 simulated long-horizon navigation episodes and identify 13 failure cases involving erroneous finish decisions, including 11 false positives (premature subtask termination and skipping) and 2 false negatives (failed subtask transitions), accounting for approximately 1.3\% of the evaluated episodes. These errors primarily arise from small residual positional deviations relative to the 1~m arrival threshold. No such errors are observed in 10 real-world long-horizon trials, although the limited sample size precludes definitive conclusions.}

    \setlength{\tabcolsep}{6pt} 
    \renewcommand{\arraystretch}{1.5}
    \begin{table}[t]
        \small
    	\centering	
    	\caption{Ablation study of the aerial navigation agent}
    	{
    		\begin{tabular}{cccc}	
    			\hline			
    			\multirow{3}{*}{\textbf{Datasets}}
                & \multicolumn{3}{c}{\textbf{Tasks}} \\
    
                \cline{2-4}
                & \multirow{2}{*}{\makecell{\textbf{Object} \\ \textbf{Navigation}}}
                & \multirow{2}{*}{\makecell{\textbf{Spatial} \\ \textbf{Reasoning}}} 
                & \multirow{2}{*}{\makecell{\textbf{Long-horizon} \\ \textbf{Navigation}}}  \\
                & & & \\ \hline
                
    			\textbf{Ours} 
    			& \textbf{98.1} & \textbf{94.6} & \textbf{88.0}  \\ \hline
                
                w/o Agent 
    			& 95.7 & 92.4 & 57.4  \\ \hline
    		\end{tabular}
    	}
    	\label{tab:agent_ablation}	
    \end{table}
    \renewcommand{\arraystretch}{1}

    \begin{figure}[t]
      \centering
      \includegraphics[width=0.82\linewidth]{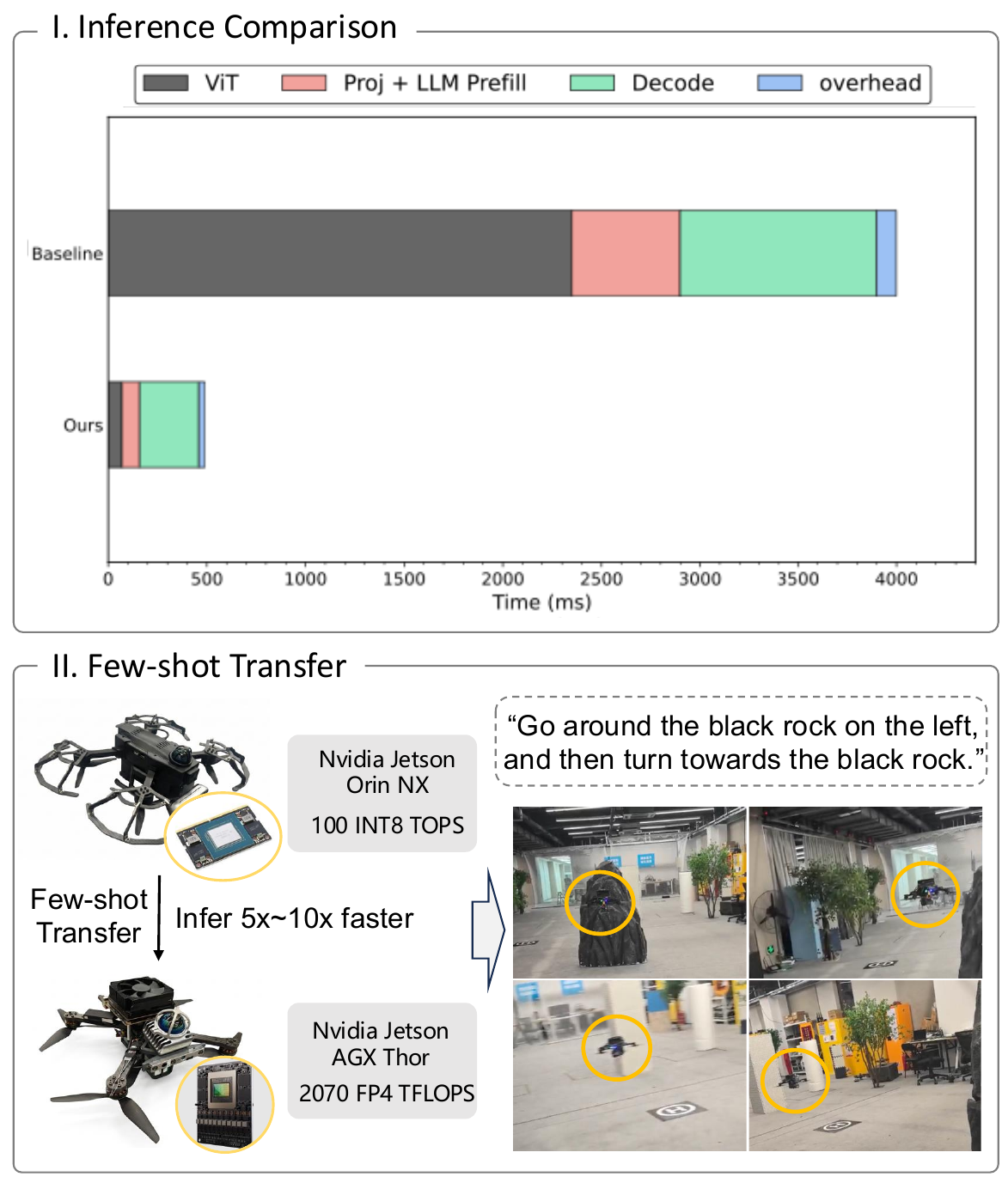}
      \caption{Ablation study of the inference optimization and the few-shot transfer.}
      \label{fig:onboard_performance}
      \vspace{-0.3cm}
    \end{figure}

    \subsubsection{Onboard Inference Optimization}
    Finally, we benchmark our optimized method on the onboard NVIDIA Jetson Orin NX platform against several popular frameworks. Specifically, we employ multiple system-level optimization strategies, including flash-attention mechanisms, KV-cache preloading, CUDA graph scheduling, and more. The performance is compared against the unoptimized baseline, and Fig.~\ref{fig:onboard_performance}(I) illustrates the inference time improvements contributed by each optimization. Notably, the total inference time is reduced from about 4100 ms to 494 ms (8.3× speedup, 87.9\% reduction). The largest absolute reduction occurs in the ViT stage (2350 → 120 ms), while both projector + LLM prefill (550 → 130 ms) and decoding (1000 → 230 ms) are substantially accelerated, together enabling real-time capable onboard deployment under constrained computational resources. In addition, we apply pipelined execution to the CPU-intensive action expert module, effectively overlapping computation and control logic to further reduce end-to-end latency on the ARM-based onboard platform. 

    Furthermore, we validate that this model inference optimization can be few-shot transferred to other hardware platforms, such as the NVIDIA Jetson AGX Thor platform\footnote{\url{https://www.nvidia.com/en-us/autonomous-machines/embedded-systems/jetson-thor/}}. To this end, we construct an aerial robot equipped with the corresponding computing platform, as shown in Fig.~\ref{fig:onboard_performance}(II), and observe similar inference acceleration benefits. In particular, compared with the NVIDIA Jetson Orin NX, the proposed system achieves 5×–10× speed improvements in onboard navigation tasks, demonstrating the strong potential of deploying large foundation models on  UAV platforms with limited computing resources.

    \section{Conclusion}
    \label{sec:conclusion}

    This paper presents VLN-AR, a hierarchical embodied aerial navigation framework that effectively decouples high-level semantic reasoning from low-level flight control for aerial robots. By leveraging a 3DGS-based automated data generation pipeline, we bridge the sim-to-real gap and address the data scarcity inherent in aerial navigation. \add{Combined with the three-stage training paradigm, the} instruction-guided token sparsification enables VLN-AR to execute sophisticated long-horizon reasoning and precise navigation on resource-constrained onboard platforms. Experimental results demonstrate superior cognitive robustness and generalization in complex, unseen environments, providing a scalable pathway for autonomous aerial robots.
    
    In future work, we aim to evolve VLN-AR into a Universal Aerial Foundation Model. Key research directions include: 1) incorporating temporal causal reasoning to navigate dynamic environments with human-robot interaction; 2) fusing multi-modal perception (e.g., point cloud and semantic map) to extend to diverse input modalities for aerial navigation under degraded sensing conditions; and 3) optimizing the framework for next-generation hardware to achieve end-to-end inference rates exceeding 10 Hz, facilitating more agile and responsive flight missions.

\section*{Acknowledgment}
The authors would like to thank Shuhang Ji, Weiqi Gai, and Jingyi Zhang for their strong support of this work.

\bibliographystyle{IEEEtran}
\bibliography{references}



\vspace{-33pt}
\begin{IEEEbiography}[{\includegraphics[width=1in,height=1.25in,clip,keepaspectratio]{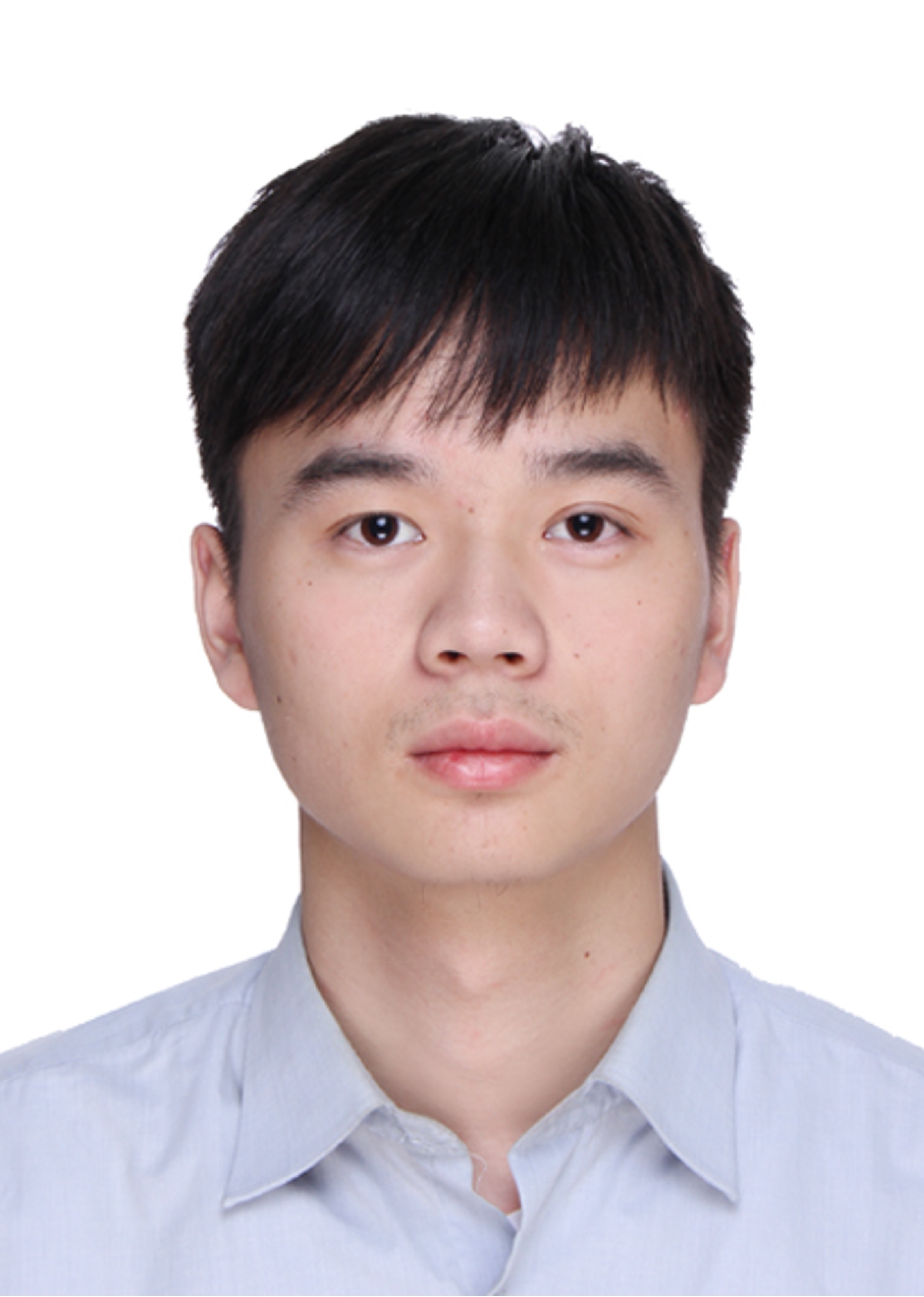}}]{Yuze Wu} received the Ph.D. degree from Zhejiang University, Hangzhou, China, in 2026. He currently leads foundation model research and development at Differential Robotics and is a joint postdoctoral researcher at Zhejiang University. His research interests include embodied intelligence for aerial robots, autonomous aerial agents, aerial manipulation, and embodied multi-robot collaboration.
\end{IEEEbiography}

\vspace{-33pt}
\begin{IEEEbiography}
[{\includegraphics[width=1in,height=1.25in,clip,keepaspectratio]{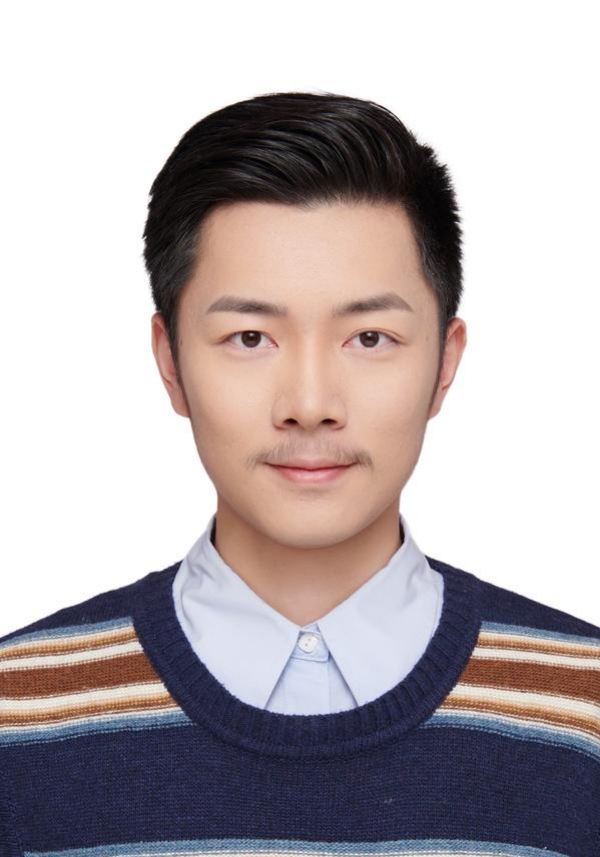}}]{Mo Zhu}
is currently pursuing the Ph.D. degree in the College of Control Science and Engineering at Zhejiang University. His research interests include large language models (LLMs) and embodied artificial intelligence (Embodied AI). His work on LLMs focuses on model compression and acceleration, particularly quantization, knowledge distillation, and speculative decoding. In Embodied AI, his research focuses on vision-language-action (VLA) models, world action models (WAMs), and autonomous unmanned aerial vehicles (UAVs).
\end{IEEEbiography}

\vspace{-33pt}
\begin{IEEEbiography}[{\includegraphics[width=1in,height=1.25in,clip,keepaspectratio]{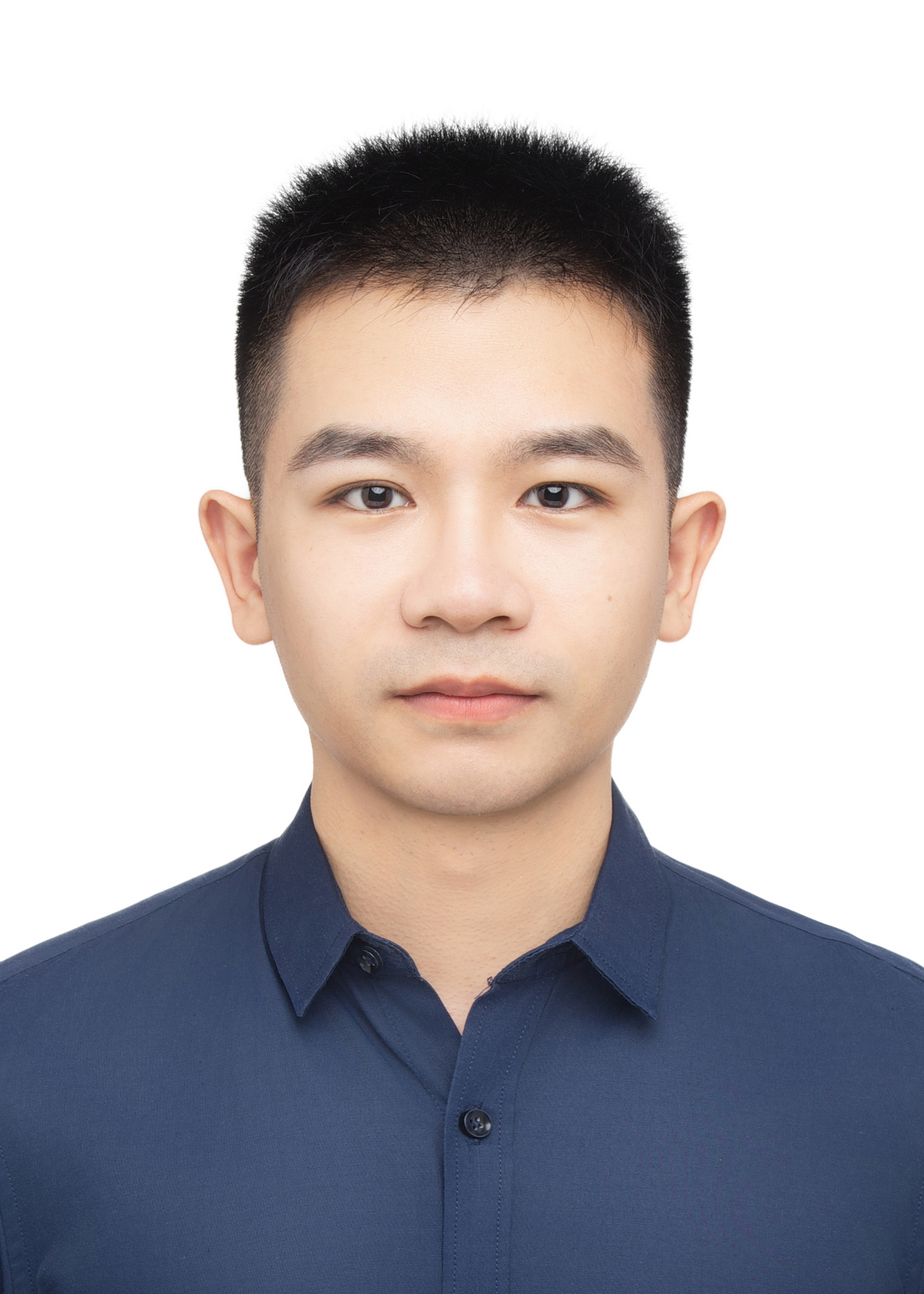}}]{Xingxing Li} received his master's degree in Data Science and Analytics from The Hong Kong University of Science and Technology (Guangzhou) in 2026 as part of the Red Bird MPhil Program (RBM). From 2019 to 2024, he worked as a Cloud Services Engineer at Huawei Consumer Business Group (CBG). He is currently an AI Algorithm Engineer at Differential Robotics (Hangzhou) Technology Co., Ltd.
\end{IEEEbiography}

\vspace{-33pt}
\begin{IEEEbiography}[{\includegraphics[width=1in,height=1.25in,clip,keepaspectratio]{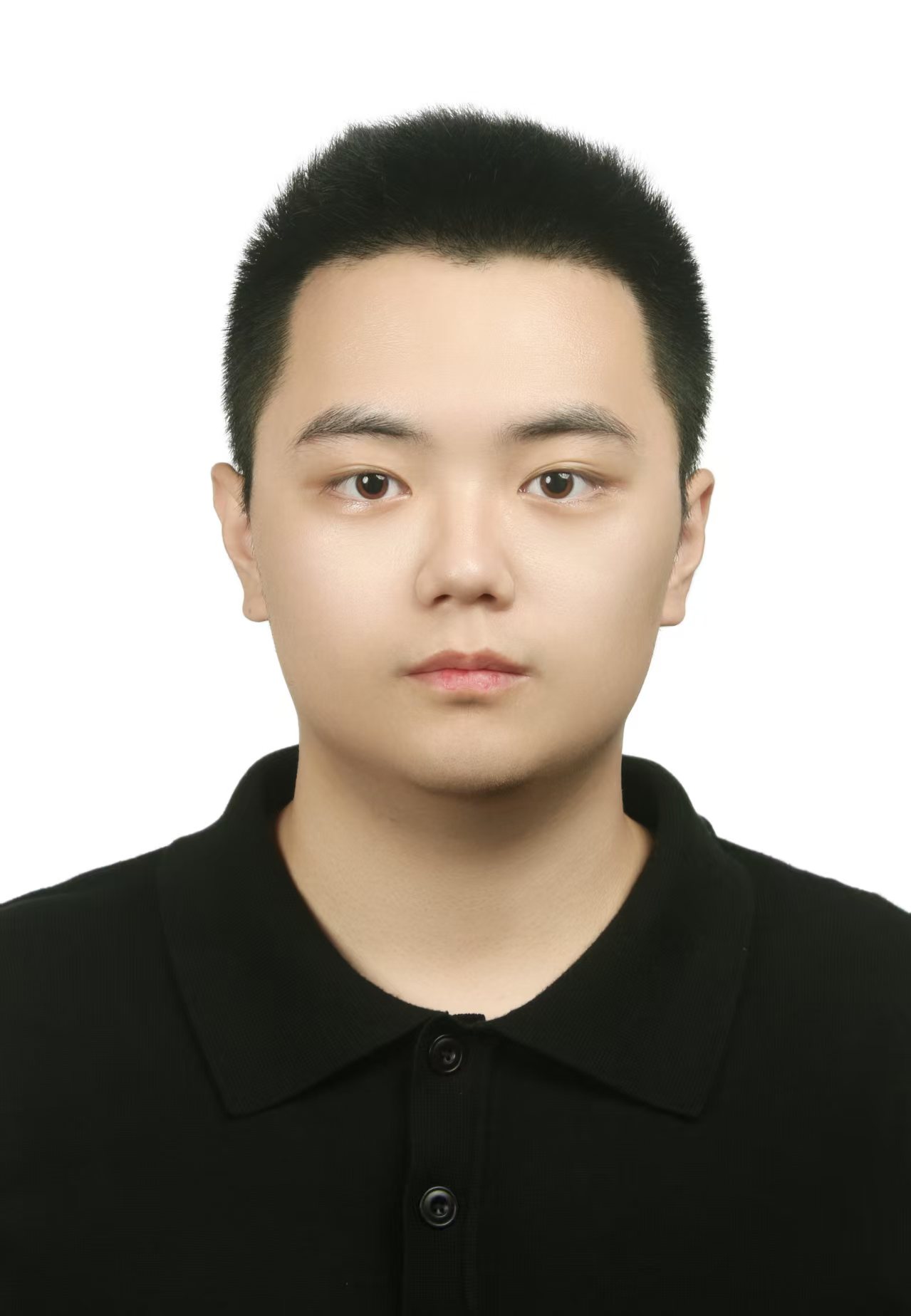}}]{Yuheng Du} received his bachelor's degree in mechanical design, manufacturing, and automation from Harbin Institute of Technology, China, in 2024, and his M.Phil. degree in Robotics and Autonomous Systems from the Hong Kong University of Science and Technology (Guangzhou), China, in 2026. He is currently with Differential Robotics (Hangzhou) Technology Co., Ltd. His research interests include unmanned aerial vehicle control and intelligent navigation systems.
\end{IEEEbiography}

\vspace{-33pt}
\begin{IEEEbiography}[{\includegraphics[width=1in,height=1.25in,clip,keepaspectratio]{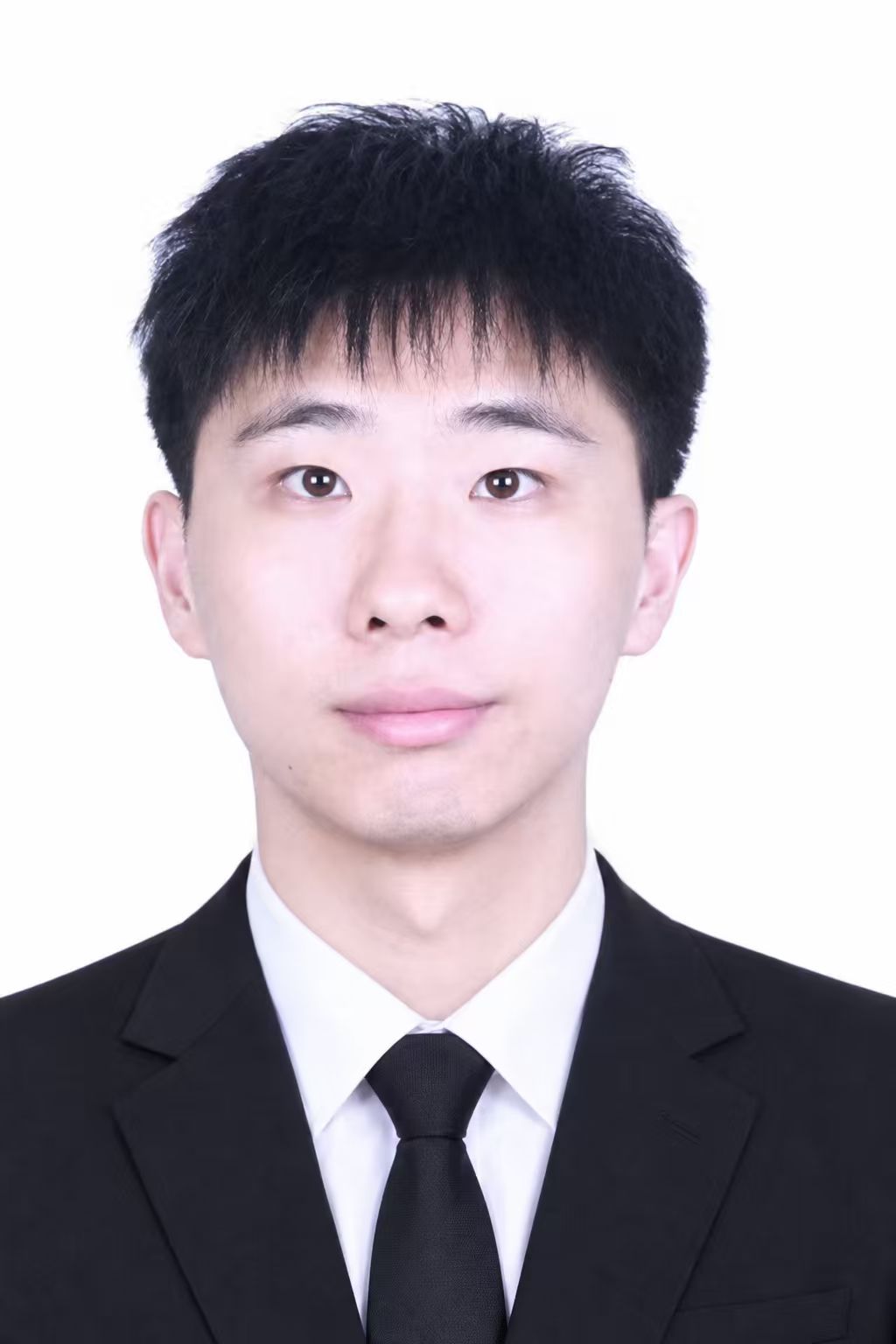}}]{Zhaoqi Wang} received his B.E. degree from Anhui University in 2022 and his M.S. degree from Beijing Institute of Technology in 2025, majoring in Control Science and Engineering. He is currently working at Differential Robotics (Hangzhou) Technology Co., Ltd. His research interests include motion planning and autonomous exploration for aerial robots.
\end{IEEEbiography}

\vspace{-29pt}
\begin{IEEEbiography}[{\includegraphics[width=1in,height=1.25in,clip,keepaspectratio]{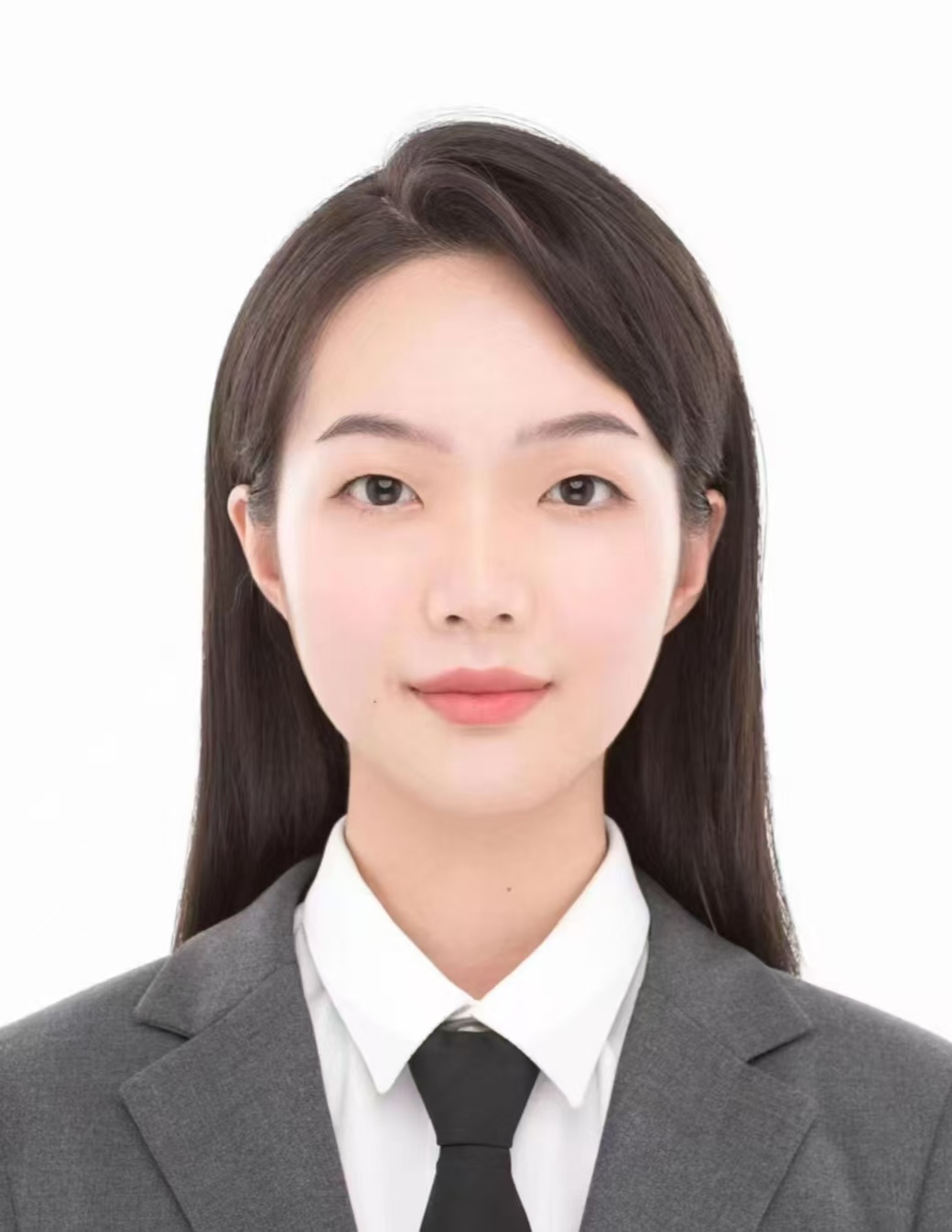}}]{Yuxin Fan} graduated from Zhejiang University in 2026, majoring in Control Science and Engineering. She is currently pursuing a master's degree in Electronic Information at the FAST Lab of Zhejiang University, under the supervision of  Fei Gao. Her research interests include path planning for multi-robot systems.
\end{IEEEbiography}

\vspace{-29pt}
\begin{IEEEbiography}[{\includegraphics[width=1in,height=1.25in,clip,keepaspectratio]{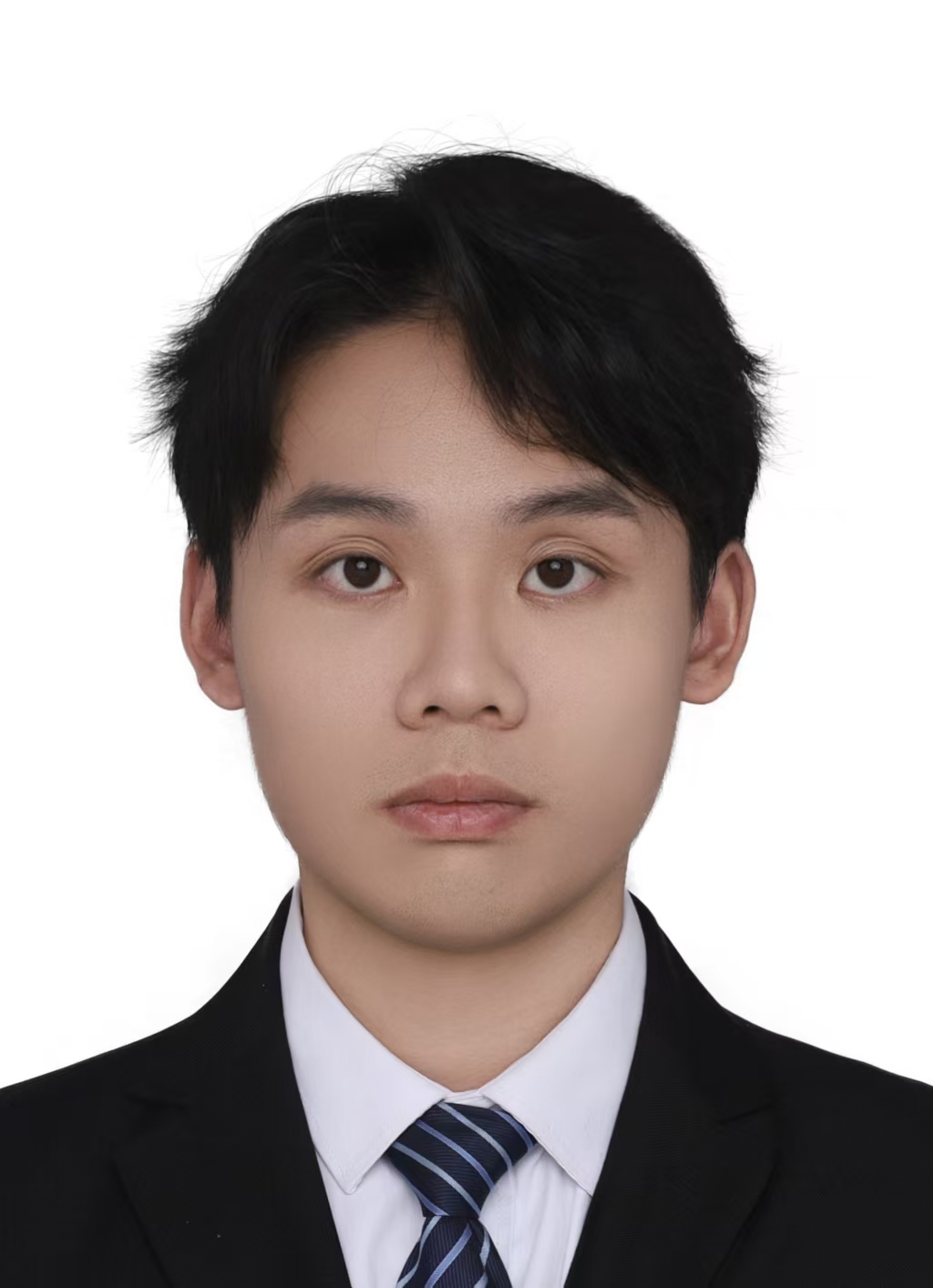}}]{Wenjun Li} received his bachelor’s degree in Control Science and Engineering from Zhejiang University in 2026. He is currently pursuing a master’s degree in Automation at Zhejiang University under the supervision of Fei Gao. His research interests include vision-language-action models and reinforcement learning.
\end{IEEEbiography}

\vspace{-29pt}
\begin{IEEEbiography}[{\includegraphics[width=1in,height=1.25in,clip,keepaspectratio]{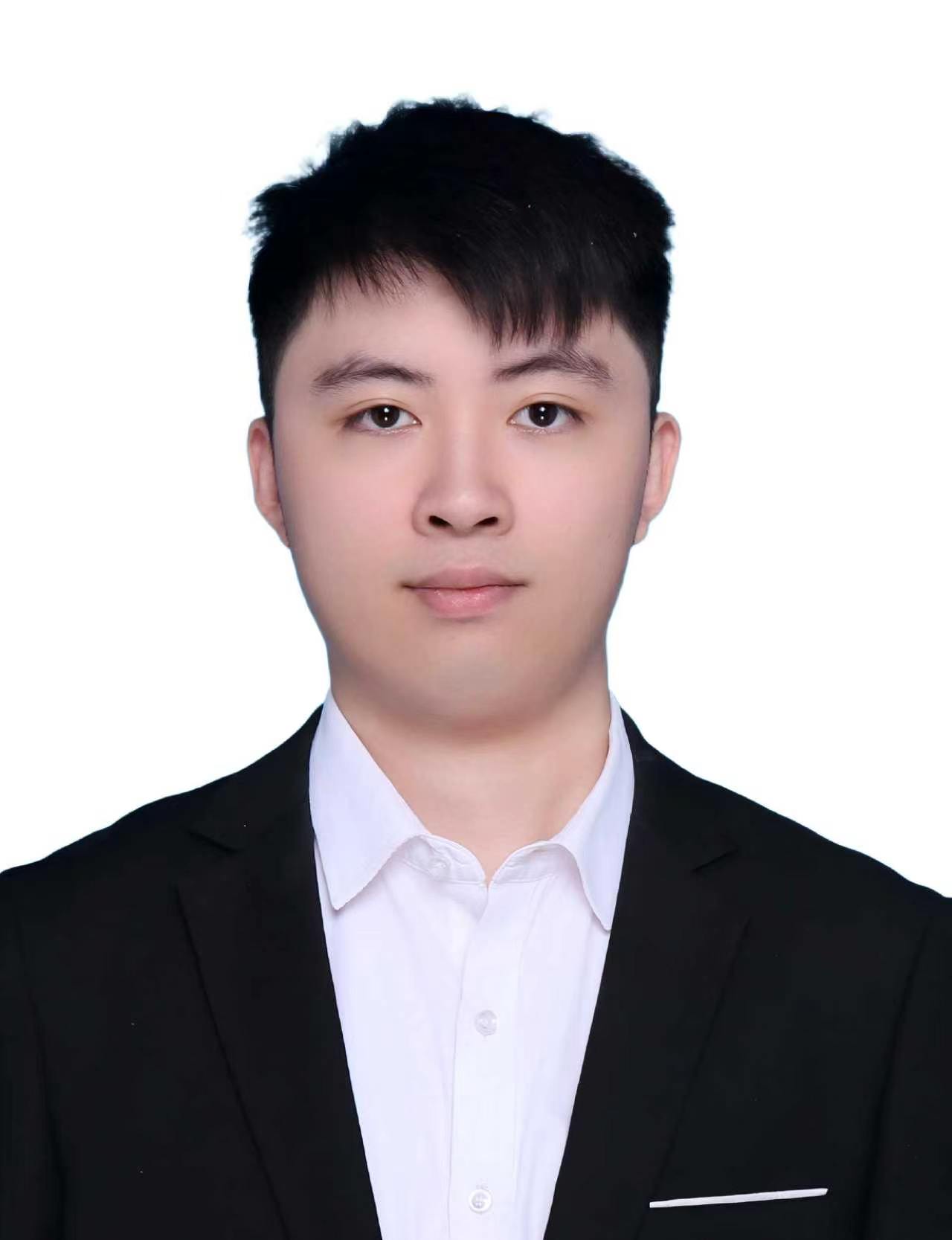}}]{Kenghou Hoi} graduated from Beijing University of Posts and Telecommunications in 2024 and is currently pursuing his master's degree at the School of Control Science and Engineering, Zhejiang University, under the supervision of Professor Fei Gao. His research interests include aerial robots, autonomous navigation and reinforcement learning for robotics. 
\end{IEEEbiography}

\vspace{-29pt}
\begin{IEEEbiography}[{\includegraphics[width=1in,height=1.25in,clip,keepaspectratio]{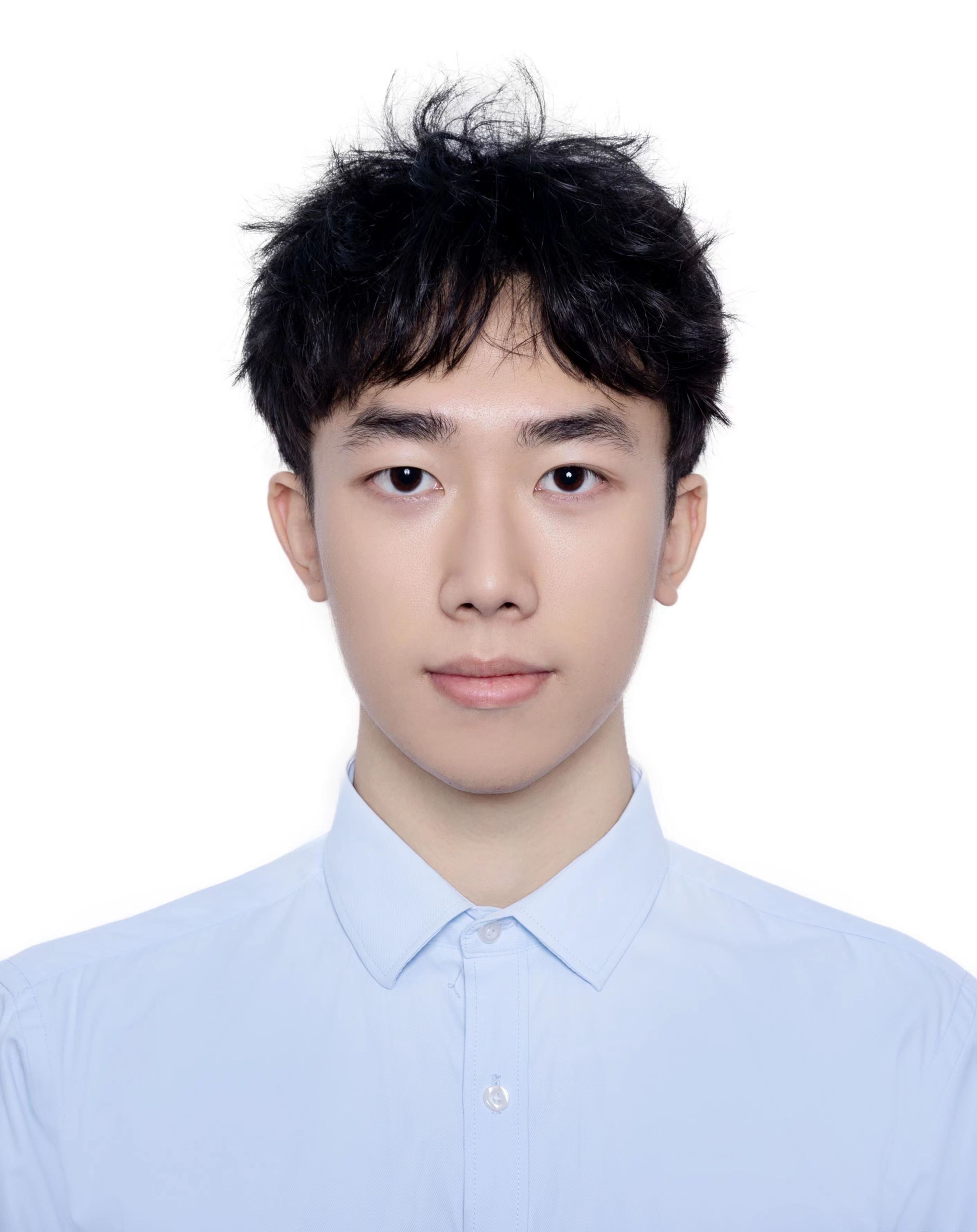}}]{Zhichao Han} received the B.Eng. degree in automation from Zhejiang University, Hangzhou, China, in 2021. He received the Ph.D. degree from Zhejiang University in 2026 under the supervision of Prof. Fei Gao at the FAST Lab. His research interests include motion planning and robot learning.
\end{IEEEbiography}

\vspace{-29pt}
\begin{IEEEbiography}[{\includegraphics[width=1in,height=1.25in,clip,keepaspectratio]{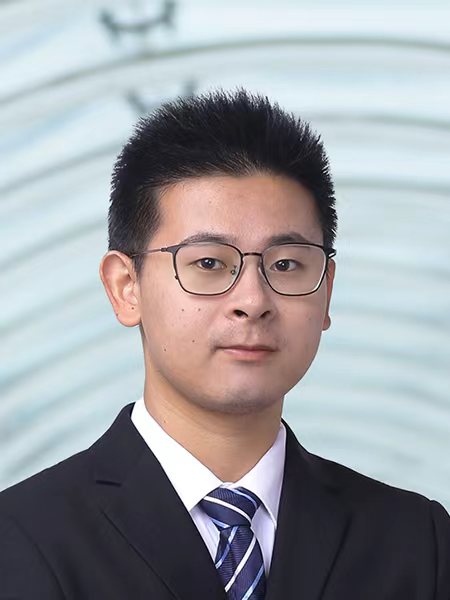}}]{Xin Zhou} received his bachelor’s degree from China University of Mining and Technology and his Ph.D. in electronic information from Zhejiang University in 2024. He subsequently conducted postdoctoral research at the Hong Kong University of Science and Technology. He is a co-founder and chief scientist of Differential Robotics. His research focuses on learning-based autonomy and swarm coordination in aerial robotics.
\end{IEEEbiography}

\vspace{-29pt}
\begin{IEEEbiography}[{\includegraphics[width=1in,height=1.25in,clip,keepaspectratio]{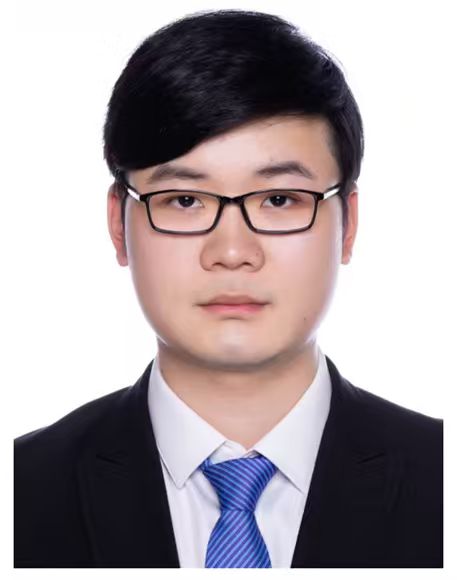}}]{Fei Gao} (Member, IEEE){\space} 
received the Ph.D. degree in electronic and computer engineering from the Hong Kong University of Science and Technology, Hong Kong, in 2019. He is currently a tenured Associate Professor in the College of Control Science and Engineering, Zhejiang University, Hangzhou, China, where he leads the Aerial Embodied Intelligence Laboratory. His research interests include aerial robots, autonomous navigation, motion planning, embodied intelligence, and swarm intelligence. He is also the founder of Differential Robotics.
\end{IEEEbiography}

\vfill

\end{document}